\documentclass[10pt,twocolumn,letterpaper]{article}

\usepackage{wacv}
\usepackage{times}
\usepackage{epsfig}
\usepackage{graphicx}
\usepackage{amsmath}
\usepackage{bbm}
\usepackage{amssymb}
\usepackage{graphicx,times,amsmath}
\usepackage{cases}
\usepackage{url}
\usepackage{multirow}
\usepackage[ruled,vlined]{algorithm2e}
\usepackage{amsfonts}
\usepackage{graphicx}
\usepackage{amssymb}
\usepackage{amsfonts}
\usepackage{subfig}
\usepackage{epstopdf}
\usepackage{makecell}
\usepackage{array}
\usepackage{xspace}
\usepackage{xparse}
\usepackage{mathtools}
\usepackage{cite}
\usepackage[symbol]{footmisc}

\makeatletter
\newcommand{\nosemic}{\renewcommand{\@endalgocfline}{\relax}}% Drop semi-colon ;
\newcommand{\dosemic}{\renewcommand{\@endalgocfline}{\algocf@endline}}% Reinstate semi-colon ;
\makeatother

\newcolumntype{?}{!{\vrule width 1pt}}

\DeclareMathOperator*{\argmin}{arg\,min}

\newtheorem{theorem}{Theorem}
\newtheorem{assumption}[theorem]{Assumption}

\usepackage[colorlinks=true,citecolor={black},breaklinks=true,letterpaper=true,bookmarks=false]{hyperref}
\makeatletter
\DeclareRobustCommand\onedot{\futurelet\@let@token\@onedot}
\def\@onedot{\ifx\@let@token.\else.\null\fi\xspace}
\def\eg{\emph{e.g}\onedot} 
\def\ie{\emph{i.e}\onedot} 
 
\def\etc{\emph{etc}\onedot}

\let\@fnsymbol\@arabic
\makeatother

\def\wacvPaperID{520} % Enter the WACV Paper ID here

\wacvfinalcopy % *** Uncomment this line for the final submission

\ifwacvfinal
\def\assignedStartPage{1} % *** Enter the assigned starting page number (instead of 9876)
\fi

\ifwacvfinal
\usepackage[breaklinks=true,bookmarks=false]{hyperref}
\else
\usepackage[pagebackref=true,breaklinks=true,colorlinks,bookmarks=false]{hyperref}
\fi

\ifwacvfinal
\else
\fi

\begin{document}

%%%%%%%%% TITLE
\title{Pushing the Envelope of Rotation Averaging for Visual SLAM
}

%\author{Xinyi Li
%Institution1 address\\
%{\tt\small xinyi.Li@temple.edu}
%\qquad 
%Haibin Ling\\
%{\tt\small hbling@temple.edu}\\
%Department of Computer Science\\
%College of Science and Technology\\
%Temple University\\
%{\tt\small \{xinyi.li, haibin.ling\}@temple.edu}
%}
\DeclareRobustCommand*{\IEEEauthorrefmark}[1]{%
  \raisebox{0pt}[0pt][0pt]{\textsuperscript{\footnotesize #1}}%
}
\author{
{Xinyi Li\IEEEauthorrefmark{1}, Lin Yuan\IEEEauthorrefmark{2}, Longin Jan Latecki\IEEEauthorrefmark{1}}
and Haibin Ling\IEEEauthorrefmark{3}\\
\IEEEauthorrefmark{1}Department of Computer and Information Sciences, Temple University, Philadelphia, USA\\
\IEEEauthorrefmark{2}Amazon Web Services, Palo Alto, CA, USA
\\
\IEEEauthorrefmark{3}Department of Computer Science, Stony Brook University, Stony Brook, USA\\
\{{\tt\small xinyi.Li, latecki@temple.edu}\}, \{{\tt\small lnyuan@amazon.com}\}, \{{\tt\small hling@cs.stonybrook.edu}\}
} 
%\vspace{-5pt}

\maketitle
%\thispagestyle{empty}
%%%%%%%%% ABSTRACT
\begin{abstract}
      As an essential part of structure from motion (SfM) and Simultaneous Localization and Mapping (SLAM) systems, motion averaging has been extensively studied in the past years and continues to attract surging research attention. 
While canonical approaches such as bundle adjustment are predominantly inherited in most of state-of-the-art SLAM systems to estimate and update the trajectory in the robot navigation, the practical implementation of bundle adjustment in SLAM systems is intrinsically limited by the high computational complexity, unreliable convergence and strict requirements of ideal initializations.
In this paper, we lift these limitations and propose a novel optimization backbone for visual SLAM systems, where we leverage rotation averaging to improve the accuracy, efficiency and robustness of conventional monocular SLAM pipelines.
In our approach, we first decouple the rotational and translational parameters in the camera rigid body transformation and convert the high-dimensional non-convex nonlinear problem into tractable linear subproblems in lower dimensions, and show that the subproblems can be solved independently with proper constraints.
We apply the scale parameter with $l_1$-norm in the pose-graph optimization to address the rotation averaging robustness against outliers. 
We further validate the global optimality of our proposed approach, revisit and address the initialization schemes, pure rotational scene handling and outlier treatments. 
We demonstrate that our approach can exhibit up to 10x faster speed with comparable accuracy against the state of the art on public benchmarks.
\end{abstract}

%%%%%%%%% BODY TEXT
\section{Introduction}
Robot navigation guided by visual information, namely, simultaneous localization and mapping (SLAM), primarily involves estimating and updating the camera trajectory dynamically. {\it Pose graph optimization}, as the fundamental element in SLAM, devotes to iteratively fix the erroneous calculation of the camera poses due to the noisy input and misplaced data association. Conventional pose graph optimization techniques are principally fulfilled with bundle adjustment (BA), which refines camera poses by progressively minimizing the point-camera re-projection errors. Fused by state-of-the-art nonlinear programming algorithms, \eg, Levenberg-Marquardt method~\cite{Levenberg1944}, Gauss-Newton method, \etc, the camera poses and map points are successively optimized according to the sequential input images.

Hindered by the intrinsic nature of visual SLAM systems, however, many challenging issues in BA remain untackled. For instance, standard BA systems yield a complexity of cubic order with regards to the input size~\cite{Wu2013b}, eventually slowing down the system to fail the real-time requirement; As the re-projection minimization involves the simultaneous optimization of considerable amounts of unknown variables, fast and accurate convergence is thus hardly achievable; Furthermore, in establishing the system stability, a good initialization scheme plays a vital role, albeit absent from most of the conventional studies.

In this work we address the aforementioned issues by leveraging multiple rotation averaging in canonical visual SLAM systems. 
First, we follow the classical $N-$point approach in epipolar geometry to solve for the relative camera rotations. Specifically, the {\it decoupling} of the rotational parameters (\ie camera orientations) and translational parameters is conducted by leveraging 6-point method, where we allocate the image features in to grids then select the feature point with the highest matching score in the grid region of the highest matching confidence.

Once we achieve the set of pairwise relative camera rotation estimations, we apply the L1-IRLS solver~\cite{chatterjee2013efficient} locally, \ie, the sub-pose-graph constructed by the frames between two consecutive keyframes. Additionally, we present an `{\it edge pruning}' process where we discard all the relative rotation estimation from last step with insufficient angular distance residual and thus guarantee the global optimality of the absolute rotation averaging solver.

With the precise estimation of the absolute camera rotations, the absolute camera rotations are considered known and fixed in the following translation averaging process. As the result of the translation ambiguity presence, we build the objective functions with additional scale parameters and equip the optimization formulation with $l_1$-norm to better handle the robustness against outliers.

By virtue of the linear, convex formulation of absolute camera translation estimation, we further validate the global optimality in all the variables of the solution given by our proposed framework. Furthermore, in the discussions we categorically address the initialization schemes and loop closure handling which are well-known to be critical yet challenging for monocular SLAM systems. We show that rotation-averaging-based SLAM systems are more robust in challenging scenarios. We conclude the paper by providing the experiment results on public benchmarks, outperformance against state-of-the-art BA-based pipelines by orders of magnitude demonstrates both efficiency and robustness of our proposed approach.

In summary, the key contributions of our paper are:
\begin{itemize}
    \item We develop a rotation averaging based optimization backbone for visual SLAM systems where camera orientations and translations are independently estimated with proper constraints.
    \item We implement additional restrictions in the rotation averaging process to ensure the global optimality, yielding faster convergence and higher accuracy.
    \item We address the challenges in monocular SLAM systems, including poor initialization, outlier handling and cumbersome loop closure, to further improve the robustness of our proposed approach.
\end{itemize}
\section{Related Work}
\begin{figure*}[th!]
\centering
\includegraphics[width=\textwidth]{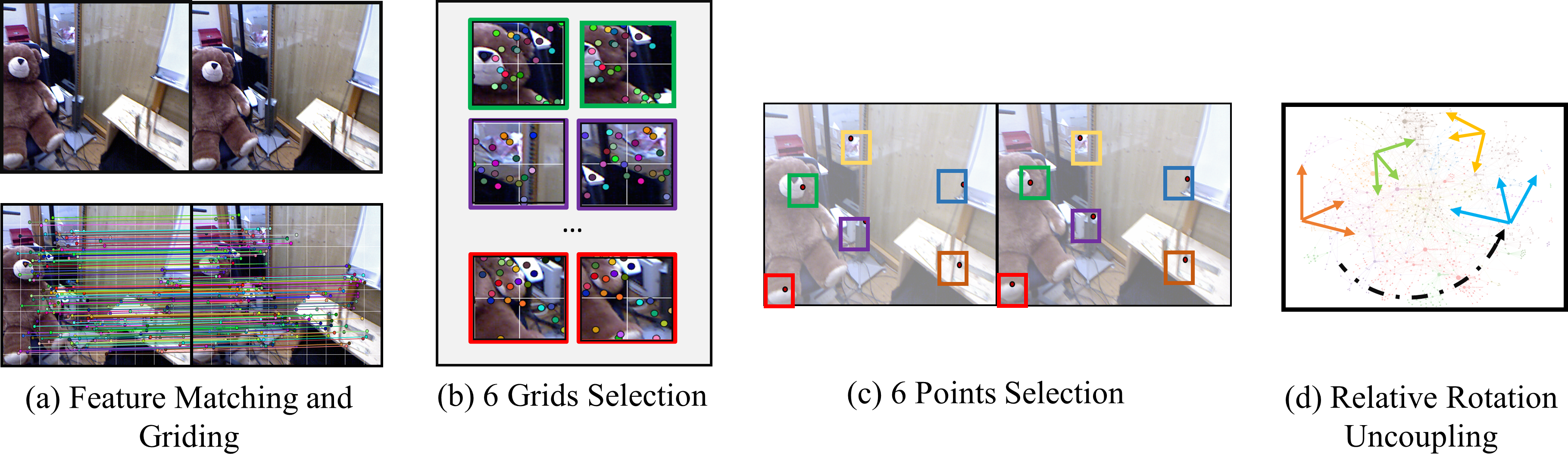}
\caption{Illustration of the decoupling of relative rotations and translations: (a) feature correspondences on the input image sequence are divided into grids, (b) grid regions with six highest matching scores are selected, these regions are normally equipped with most features, (c) points with highest matching confidence in each grid are selected, (d) the relative rotation is solved by iteratively choosing two pairs among them. Details are presented in \S\ref{sec:relrot}.}
\label{fig:pipeline}
\end{figure*}
Camera pose estimation lies in the heart of monocular SLAM systems, whereas the camera orientation and translation optimization consist the camera pose refinement process. Compared with conventional BA-based approaches where re-projection errors are iteratively minimized, approaches fused by rotation averaging methods have been recently proven more efficient yielding comparable or higher accuracy which greatly benefit real-time applications with limited computational power. Rotation averaging has been first introduced into 3D vision by~\cite{govindu2001combining} where the authors exploit Lie-algebraic averaging and propose an efficient and robust solver for large-scale rotation averaging problems, and was later studied in~\cite{moakher2002means}. Outperformance with the motion averaging backbone against canonical approaches stimulates numerous SfM frameworks~\cite{cui2015, zhu2018, zhu2017parallel, cui2017hsfm, Li20Hybrid, martinec2007robust, locherprogsfm}, whereas global motion averaging is conducted to simultaneously solve all camera orientations from inter-camera relative motions. In~\cite{cui2017hsfm}, the authors develop a camera clustering algorithm and present a hybrid pipeline applying the parallel-processed local increment into global motion averaging framework. Similarly, in~\cite{zhu2018}, distributed large-scale motion averaging is addressed. In~\cite{Li20Hybrid}, a hybrid camera estimation pipeline is proposed where the dense data association introduces a single rotation averaging scheme into visual SLAM.

Rotation averaging~\cite{hartley2013rotation} has shown improved robustness compared with canonical BA-based approaches in numerous aspects. For instance, proper initialization plays a vital role in equipping a sufficiently stable monocular system ~\cite{tang2017gslam}, while~\cite{carlone2015initialization} addresses the initialization problem for 3D pose graph optimization and survey 3D rotation estimation techniques, where the proposed initialization demonstrates superior noise resilience. In~\cite{chatterjee2013efficient} and~\cite{chatterjee2018robust} the process is initialized by optimizing a $l_1$ loss function to guarantee a reasonable initial estimate.  It has also been shown in 567 that estimating rotations separately and initialize the 2D pose graph with the measurements provide improved accuracy and higher robustness. In\cite{kneip2012finding, kneip2014efficient} it has been exploited that camera rotation can be computed independent of translation given specific epipolar constraints. It is well known that monocular SLAM is sensitive to outliers and many robust approaches~\cite{govindu2006robustness, yang2020graduated, enqvist2011non, kahl2008multiple} have thus been designed to better handle the noisy measurements. Moreover, Lagrangian duality has been reconciled in recent literature~\cite{fredriksson2012simultaneous, briales2018certifiably, carlone2015lagrangian} to address the solution optimality. A recent paper~\cite{eriksson2018rotation} shows that certifiably global optimality is obtainable by utilizing Lagrangian duality to handle the quadratic non-convex rotation constrains~\cite{wilson2016rotations} and further derives the analytical error bound in the rotation averaging framework.  

Our proposed approach has been inspired by recent work~\cite{bustos2019} and~\cite{Li20Hybrid}. In~\cite{Li20Hybrid}, the authors partition the input sequence into blocks according to the pairwise covisibility and the optimization is processed hierarchically with local BA and global single rotation averaging. While~\cite{Li20Hybrid} yields high accuracy, it is demanding to handle the latency between local and global optimization and the system may suffer time overhead progressively.
In~\cite{bustos2019}, the camera orientation estimation is incrementally conducted, followed by a quasi-convex formulation to solve for the camera and map point positions in the known rotation problem, namely, {\it L-infinity SLAM}. While the authors of~\cite{bustos2019} state that L-infinity SLAM is not globally optimal, we modify the relative rotation estimation process to yield the theoretical globally optimal solution. Moreover, L-infinity SLAM formulates the camera-point optimization with $l_{\infty}$-norm of the re-projection errors, which is susceptible in two-folds: 1) The optimization in $l_{\infty}$-norm is well-known to be sensitive to outliers, especially for the noisy nature of SLAM problems; 2) Similar with the conventional BA approaches, the size of the KRot problem~\cite{zhang2018fast} is not tractable and keeps increasing in dimension as the process continues. In contrast, we incorporate the linear global translation estimation constraints proposed in~\cite{cui2015linear} in our translation averaging. Note that no scene point positions are required such that our framework is robust in dealing with large-scale data. 

\section{Preliminaries and notations}
Consider a 3D scene with $m$ frames with known camera intrinsic matrices $K_1, \dots, K_m$, where $K_i \in \mathcal{K} \subseteq \mathbb{R}^3, 1 \leq \forall i \leq m$ is known and fixed. Let $\mathcal{G} = \{\mathcal{V}, \mathcal{E}\}$ denote the pose graph for the given scene, where the vertices in $\mathcal{V}$ represent the absolute camera rotations and the edges in $\mathcal{E}$ denote that the relative motion between the two camera rotations can be recovered, \ie, $e_{ij} \in \mathcal{E}$ if the data association between $i^{\text th}$ frame and $j^{\text th}$ frame is adequate to estimate the corresponding camera relative motion.

Formally, denote $C_i = [\mathbf{R}_i | t_i] \in \mathcal{C} \subseteq \mathbb{SE}(3) \subseteq \mathbb{R}^{3 \times 4}$ for the absolute camera pose for the $i^{\text th}$ frame, where $R_i \in \mathbb{SO}(3) \subseteq \mathbb{R}^{3 \times 3}$ denotes the absolute camera rotation and $t_i \in \mathbb{R}^3$ denotes the absolute camera translation in the global coordinate system. By convention, we define the essential matrix $E_i = [t_i]_\times R_i$, where
\begin{equation}
    [t_i]_\times = 
    \begin{pmatrix}
    t_1^i\\
    t_2^i\\
    t_3^i
    \end{pmatrix}
    =
    \begin{pmatrix}
    0 &-t_3^i& t_2^i\\
    t_3^i & 0 & -t_1^i\\
    -t_2^i & t_1^i & 0
    \end{pmatrix}.
\end{equation}

The relative rotation from frame $i$ to frame $j$ is denoted by $R_{ij}$, specifically
\begin{equation}
    \label{eq:reldef}
    \mathbf{R}_{ij} = \mathbf{R}_j \mathbf{R}_i^\top, 1 \leq i,j \leq m.
\end{equation}

Assume that we can estimate a set of relative camera rotations and denote the corresponding measurements between frame $i$ and $j$ as $\widetilde{\mathbf{R}}_{ij}$, we thus define the objective function of rotation averaging as
\begin{equation}
    \label{eq:rotavgdef}
    \argmin_{\mathbf{R}_{ij} \in \mathbb{SO}(3)}\sum_{1 \leq i,j \leq m} \mathbbm{1}_{ij}d(\mathbf{R}_{ij}, \widetilde{\mathbf{R}}_{ij}),
\end{equation}
where $\mathbbm{1}_{ij}$ is the abbreviation of $\mathbbm{1}_{e_{ij} \in \mathcal{E}}$ and $d(\mathbf{R}_{ij}, \widetilde{\mathbf{R}}_{ij})$ denotes the estimation error. Throughout the paper we use the chordal distance for the metric of $d(\cdot, \cdot)$, where this commonly used metric is defined as the Euclidean distance between the two rotations on the embeddings in $\mathbb{R}^9$. Specifically,
\begin{equation}
    \label{eq:chordal}
    d(\mathbf{R}_{ij}, \widetilde{\mathbf{R}}_{ij}) = \|\mathbf{R}_{ij} - \widetilde{\mathbf{R}}_{ij}\|_{F},
\end{equation}
where $\|\cdot\|_{F}$ denotes the Frobenius norm. In addition, we denote the angular distance $\alpha _{ij}$ between $\mathbf{R}_i$ and $\mathbf{R}_j$, defined as 
\begin{equation}
    \label{eq:angdef}
    \alpha _{ij} = \angle (\mathbf{R}_i\widetilde{\mathbf{R}}_{ij}, \mathbf{R}_j),
\end{equation}
and we also have the following equivalency 
\begin{equation}
    \label{eq:angular}
    d(\mathbf{R}_{ij}, \widetilde{\mathbf{R}}_{ij}) = 2\sqrt{2}\sin (\alpha _{ij} / 2).
\end{equation}
We will use the consistent metric in $l_1$-norm throughout the paper, further implementation details are illustrated in~\ref{sec:approach}.

\section{Our approach}
\label{sec:approach}
We aim to solve for the absolute camera poses and refine the computational results with progressive data input for visual SLAM systems. In this section, we describe our algorithm by starting with the relative rotation-translation decoupling in \S\ref{sec:relrot}, followed by the epipolar geometry based rotation estimation details. An illustration of the relative rotation estimation process is given in Fig.~\ref{fig:pipeline} Then we depict the absolute rotation solving steps with the `edge pruning' process derived from the global optimality assumption in \S\ref{sec:rotavg}. Once the global rotations are achieved, translation averaging is solved in a linear manner with scale awareness, detailed in \S\ref{sec:transavg}. 
\subsection{Relative Rotation Estimation}
\label{sec:relrot}
It has been well proven that enforcing additional epipolar coplanarity constraints can guarantee the safe decoupling of the camera rotation and translation computations~\cite{kneip2012finding, kneip2014efficient}. However, the $N-$point method cannot be directly fused into SLAM pipeline due to its high computational complexity. In general, it requires $n \choose 2$ combinations of $n$ feature correspondences, which is prohibitive in practical implementations. In our approach, we instead select the six feature correspondences which are the most `computationally representative', \ie, the six feature pairs are those with highest confidence matching score within the region of the most dense features. As we will show in \S\ref{sec:exp}, though it takes slightly more time for the feature confidence sorting compared to exhaustive brute-force feature matching, the performance of our proposed approach is superior to BA-based methods on the efficiency especially for scenes with rich features. 

\begin{figure}[t!]
\centering
\includegraphics[width=\linewidth]{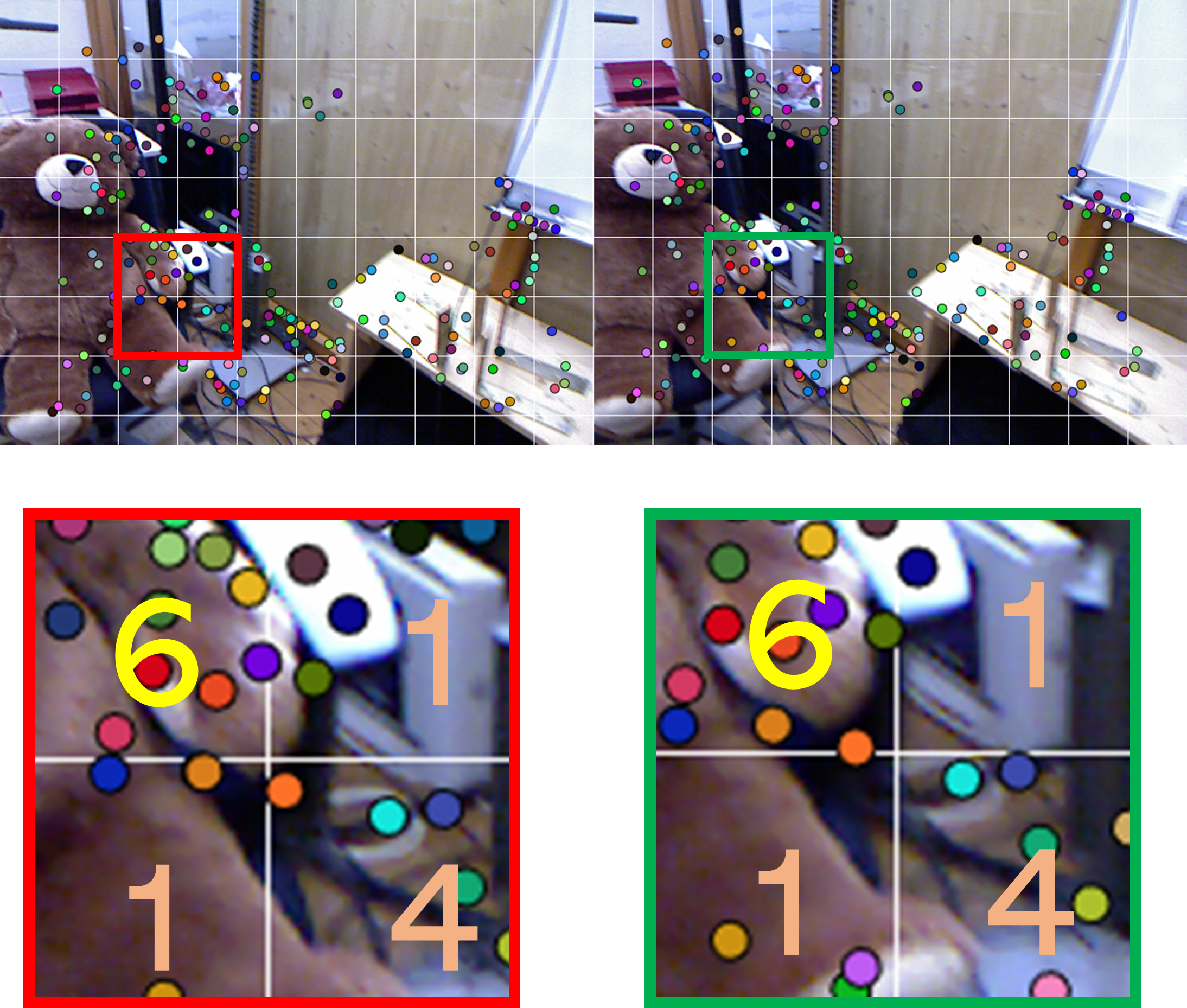}
\caption{Illustration on the feature region correspondence score computing. The two sub-grids are taken from the same location from the original image pairs, only the matched features in the corresponding are counted towards the region correspondence score.}
\label{fig:grid}
\end{figure}

In details, after processing feature extraction on the input images, we first divide the input images into $n$ grids of the same size, and sort the grids according to the `{\it correspondence score}'. Note that the correspondence score is not the amount of the feature points in the grid, as we only count the matched features in the corresponding grids. For example, in Fig.~\ref{fig:grid}, we take two corresponding $2 \times 2$ sub-grids of the matched features, for grid (red) $(1,1)$ we only sum up the matched features which also appear in the corresponding grid (green) $(1,1)$, which is $6$. In this way, we achieve the feature regions with the most dense valid feature correspondences. Practically, an even 
scattering of the six regions is desired as it can reserve most information of the images. Once the grid selection is accomplished, one feature correspondence pair with the highest feature matching confidence score is picked in each region. It is well-known that relative rotation can be solved by $N-$point method where $N\geq 5$, we use $6$ points to achieve the balance between accuracy and speed.

Now given $6$ feature correspondences $(p_n^i, p_n^j), 1\leq n \leq 6$, we iteratively select two of them to compute the relative rotation and only keep those where the relative translation is sufficiently small for the epipolar constrains, followed by the least squares problem 
\begin{equation}
    \argmin _E \sum_{n = 1}^6 \|{f_n^i} ^\top E f_n^j\|_2^2,
\end{equation}
where $f_n$ denotes the unit feature vector, \ie, $f_n^i = \dfrac{p_n^i}{\|p_n^i\|}$. To ensure that the rotation and translation can be independently solved, we require the following constraint.
\begin{assumption}
\label{thm:assump}
Let $f_n, \widetilde{\mathbf{R}_{ij}}$ be as defined above, the rotation can be solved independent of translation iff. 
\begin{equation}
    |(f_n^{i_1} \times \widetilde{\mathbf{R}_{ij}}) (f_n^{j_1} f_n^{i_1} \times \widetilde{\mathbf{R}_{ij}}) (f_n^{j_1} f_n^{i_1} \times \widetilde{\mathbf{R}_{ij}} f_n^{j_1})|=0,
\end{equation}
$1 \leq \forall i^\cdot, j^\cdot \leq 6$.
\end{assumption}
The proof to Assumption~\ref{thm:assump} is beyond the scope of this study, further details including the theoretical solution can be found in~\cite{kneip2012finding}.

In our approach, relative rotation estimation is incrementally conducted along with incoming image frames. For any two consecutive frames, we solve for the relative rotation as described above. A brief illustration is given in Fig.~\ref{fig:pipeline}.

Note that our approach is processed within a keyframed pipeline, \ie, a keyframe is marked every 30 frames and the relative rotation estimation is only incrementally conducted locally. In such way, we can 1) avoid accumulated errors and 2) effectively control the size of sub-problems to ensure the real-time performance.
\subsection{Absolute Rotation Estimation}
\label{sec:rotavg}
\begin{figure}[t!]
\centering
\includegraphics[width=\linewidth]{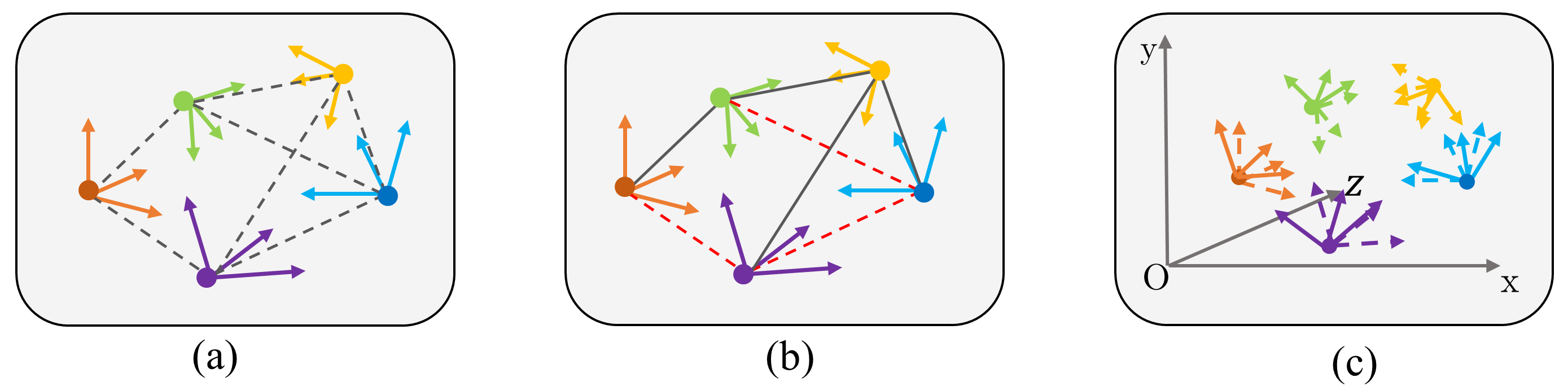}
\caption{Our absolute rotation estimation~\ref{sec:rotavg} process takes pose graph as input where the edges represent the relative rotation estimations~\ref{sec:relrot}. (a): the camera orientations in the initialization with the computed pairwise relative rotations (dashed lines) connecting them. (b) During each L1-IRLS run, those edges not satisfying the angular error bound condition (dashed lines) are removed and isolated from the optimization. The solid lines represent the edges which are kept. (c): Once the absolute rotation estimation is finished, the orientations (solid lines) are adjusted according to the world coordinate.}
\label{fig:relrot}
\end{figure}

Once we achieve the pairwise relative rotation estimation, we proceed to average the results to solve for the absolute camera rotations. In our approach we adopt L1-IRLS solver~\cite{chatterjee2013efficient} for Eq.~\ref{eq:rotavgdef}. Additionally, we notice that in practical use the L1-IRLS is likely to stuck at local minima when the pairwise relative rotation is sufficiently small, which bares a high angular error. We thus implement an {\it angular error bound} into the iteration, such that erroneous edges in the partial pose graph can be effectively removed. This `{\it pruning process}' can also theoretically guarantee the global optimality (further discussed in \ref{sec:globalopt}), while the relaxation accelerates the convergence. The illustration is given in Fig.~\ref{fig:relrot} and we will discuss the details in the following.

As the absolute rotation averaging is conducted on each sub-pose-graph of 30 local frames which share plentiful covisible map points, camera poses are normally highly connected due to the dense data association. Generally, dense data correlation bring high amounts of noise into the pose-graph optimization. We have explored that both the accuracy and the efficiency deteriorate with too many edges. The edge pruning is embedded into the L1-IRLS steps and conducted in such a way that, each time when we achieve an optimization result from the L1-IRLS solver, we check the angular distance residual as defined in Eq.~\ref{eq:angdef} between all local rotation pairs. The theoretical angular threshold~\cite{eriksson2018rotation} $\alpha_{\max}$ is given by
\begin{equation}
    \alpha _{\max} = 2\arcsin\bigg(\sqrt{\frac{1}{4}+ \frac{\lambda_2(L_G)}{2d_{\max}}}-\frac{1}{2}\bigg),
\end{equation}
where $\lambda _2 (L_G)$ denotes the Fiedler value~\cite{fiedler1973algebraic} of the pose graph $G$ and $d_{max}$ denotes the maximal vertex degree. We set $\alpha _{\max} = 45^{\circ}$ for practical use. For the rotation pairs with angular residual higher than the threshold, we replace the relative rotation measurement with the `optimal relative rotation'. For example, assume there exist $\mathbf{R}_i ^*$ and $\mathbf{R}_j ^*$ (where $\mathbf{R}^* _{\boldsymbol{\cdot}}$ denotes the IRLS computational result), such that
\begin{equation}
    \label{eq:angthresh}
    \alpha_ {ij} =\angle(\mathbf{R}_i ^* \widetilde{\mathbf{R}}_{ij}, \mathbf{R}_j ^*)>\alpha _{\max},
\end{equation}
we replace $\widetilde{\mathbf{R}}_{ij}$ with the `pseudo' optimal relative rotation $\mathbf{R}_{ij} ^*$, where $\mathbf{R}_{ij} ^* = \mathbf{R}_j ^* {\mathbf{R}_i ^{*\top}}$. By resetting the value of this edge, it is equivalent as removing the constraints of this erroneous relative rotation from further optimization process. The described process is repeated with every IRLS run, in practice, we set the maximum number of edge pruning iterations to be 3 to compromise speed and accuracy. The illustration of the rotation estimation is provided in Fig.~\ref{fig:relrot}.
\subsection{Translation Averaging}

\label{sec:transavg}
With the absolute camera rotations recovered from the previous steps, we hereafter treat them as fixed and known parameters in calculating the absolute translations of the camera motion. As only the directional relative translation (unit vector) can be directly recovered from the essential matrices, we continue to solve for the absolute translations  $x = \{t_i| t_i \in \mathbb{R}^3, \forall i\}$ by conducting translation averaging with unknown scale parameters $\{s_i| s_i \in \mathbb{R_{+}}, \forall i\}$. 

Following~\cite{cui2015linear}, first assuming a unit baseline length between camera $i$ and camera $j$, we can solve $s_i$ and $s_j$ as below
\begin{equation}
    \label{eq:scale}
    (\mathbf{R}_i t_{ij} \times (-\mathbf{R}_j)t_{ij}) \times [(t_i - t_j) + \frac{1}{s_i}\mathbf{R}_i t_{ij} + \frac{1}{s_j}\mathbf{R}_j t_{ij}] = 0,
\end{equation}
here `$\times$' denotes the cross product operation. Let $\mathbf{A}_i^{ij} = (s_i \|t_j - t_i\|\mathbf{R}_i)^{-1}$ and $\mathbf{A}_j^{ij} = (s_j \|t_j - t_i\|\mathbf{R}_j)^{-1}$, we can further generalize the system as 
\begin{equation}
    \label{eq:multconstraint}
    (\mathbf{A}_j^{ij} - \mathbf{A}_i^{ij})(t_i - t_j) + t_i + t_j = (\mathbf{A}_l^{kl} - \mathbf{A}_k^{kl})(t_k - t_l) + t_k + t_l,
\end{equation}
where $1 \leq i\neq j\neq k \neq l \leq m$, \ie, we can expand the system with more camera pairs of known rotations as long as there exist sufficient inter-pair covisibility. Denote $\mathbf{A}$ for the coefficient matrix associated with the final system of stacked Eq.~\ref{eq:multconstraint}, and denote $x$ as the concatenated stacked vector of absolute camera translations. Then the objective function is equivalent as
\begin{equation}
\label{eq:transopt}
\argmin_{x \in \mathbb{\mathbf{R}}^{3m}}\|\mathbf{A}x\|_1,{\hspace{5pt}\text s.t.\hspace{5mm}} \|x\|_2 = 1.
\end{equation}
Equivalently, letting $e = \mathbf{A}x$, the augmented Lagrangian function of Eq.~\ref{eq:transopt} is
\begin{equation}
\label{eq:lagran}
    L(e, x, \lambda) = \|e\|_1 + \langle \lambda, \mathbf{A}x - e \rangle + \frac{\beta}{2}\|\mathbf{A}x - e\|_2^2,{\hspace{5pt}\text s.t.\hspace{5mm}} \|x\|_2 = 1,
\end{equation}
where $\lambda$ is the Lagrange multiplier, $\langle\cdot, \cdot\rangle$ denotes the inner product operation, and $\beta$ is the relaxation parameter.
Then the ADMM iteration is defined as below
\begin{subequations}
\begin{align}
    e_{k+1} &= \argmin_e \|e\|_1 + \langle \lambda_k, \mathbf{A}x_k - e \rangle + \frac{\beta}{2}\|\mathbf{A}x_k - e\|_2^2, \\
    x_{k+1} &= \argmin _{\|x\|_2 = 1} \langle \lambda_k, \mathbf{A}x_k - e_{k+1} \rangle + \frac{\beta}{2}\|\mathbf{A}x - e_{k+1}\|_2^2,\\
    \lambda_{k+1} &= \lambda_k + \beta(\mathbf{A}x_{k+1} - e_{k+1}).
\end{align}
\end{subequations}
The absolute camera translation can thus be recovered from the stationary point $x^*$ of the system. Note that instead of explicitly using map point locations, this method only relies on the geometric constraints which is more robust compared to the counterparts, especially when dealing with large-scale data.

%To tackle the scale ambiguity, we set $s_1 = 1$ and set the camera rotation $R_i = I$ to address the gauge freedom.
\section{Discussions}
\subsection{Global Optimality}
\label{sec:globalopt}
We first show the global optimality of the rotation averaging estimation, we will restate the condition (also known as strong duality condition) as below. 
\begin{theorem}
\label{thm:duality}
Consider the objective function defined in Eq.~\ref{eq:rotavgdef} for a connected pose graph, let $\mathbf{R}_i^*$ be a stationary point, then $\mathbf{R}_i^*, i=1, \cdots, n$ are globally optimal if $|\alpha _{ij}| \leq \alpha _{\max}$.
\end{theorem}
The full proof to Thm.~\ref{thm:duality} is given in~\cite{eriksson2018rotation}. 
In our approach, as previously discussed in \ref{sec:rotavg}, we impose the angular error bound into the rotation averaging process such that the strong Laplacian duality holds for all post-pruning edges in the pose graph. Furthermore, we address the connectedness condition by conducting absolute rotation estimation on the sub-graphs where all camera viewpoints are close to each other, so they are mostly connected.

Since in our approach, we construct the absolute translation averaging in $l_1$-norm, the formulation is thus convex and it immediately follows that all the stationary points are global optimal solution~\cite{candes2005decoding}. Therefore the proof is trivial and we claim that our approach can achieve global optimality in all unknown variables.
\subsection{Robustness}
\label{sec:robust}
\subsubsection{Initialization}
\label{sec:init}
A good initialization is critical for the robustness of the SLAM system and many visual SLAM systems suffer from slow/cumbersome initialization. In the typical pipeline of BA-based approaches, successful initialization rely heavily on the covisibility graph, \ie, the system cannot initialize without the presence of sufficient map points. In an environment lack of visible features, tracking tends to get lost soon even if the system initializes. By contrast, our approach requires very few covisible map points (theoretically the system can initialize with six good feature points) and the initialization takes much shorter time.

Furthermore on a numerical scope, as proposed in~\cite{chatterjee2018robust}, the rotation averaging is initialized with $l_1$ optimization, the output of which enters the second step as the initial guess to ensure a starting point sufficiently close to the optimal solution. In practice, however, we have observed that it is unnecessary to fully finish the iterations of the `L1RA' step to provide a good initial value. Indeed, the original L1-IRLS algorithm cannot fulfill the real-time requirement. In our experiments, we have set a threshold for the maximum iteration number for L1RA steps. 

In addition, initial values of the translation can be arbitrary since we adopt the linear solver in translation estimation, the initialization is further simplified compared with traditional SLAM pipelines.

\subsection{De-noising and Constraints}
\label{sec:constraints}
Compared with BA based methods, approaches integrated with rotation averaging are proven to be generally more robust against noisy data. In our approach, we address the outlier rejection in several aspects to further enhance the robustness to handle scenarios of different settings or scales. During the relative rotation estimation, we actively limit the feature pair selection into a small, reliable region, instead of the random selection proposed in~\cite{kneip2012finding}. In addition, we add the edge pruning in the rotation averaging process to minimize the possible negative effects from mismatched features. The translation averaging process in our proposed approach is formulated in $l_1$-norm, known as far less sensitive to outliers~\cite{wilson2014robust} than $l_2$ or $l_{\infty}$ estimations. We can thus claim that our approach is highly robust, as demonstrated in the experiments.
\subsection{Loop Detection and Closure}
\label{sec:loopclos}
In contrast with the conventional loop detection and closure schemes where loop closure is triggered when sufficient feature matching inliers are detected, our approach observes loop according to two-step criterion: 1) Two absolute camera translations are close enough to each other, \ie, the Euclidean distance of the two cameras is smaller than a threshold; 2) Sufficient feature inliers are observed. Updating the complete covisibility graph in a high frequency for rotation averaging based approaches is redundant and is avoided in our approach. 

In detail, if a loop is detected at camera $s$ with a previous camera position $t$, we mark both frames as keyframes and conduct the translation averaging process on all the interval keyframes between these two frames. Instead of calculating the re-projection errors, we set absolute camera rotations and the translation scale parameters as fixed and known to reproduce the augmented Lagrangian function as in Eq.~\ref{eq:lagran}. Once the loop closure is finished, the adjustments on the absolute translations are propagated to all the frames associated with the keyframes according to the relative translations. Note that the loop closure process can be extended to include the optimization of the rotations in the similar manner to achieve better precision, however, it will take significantly longer processing time and be only suitable for offline approaches.  

\section{Experimental Results}
\label{sec:exp}
\subsection{Implementation Details}

{\it System Configuration} We conduct all of our experiments on a PC with Intel(R) i7-7700 3.6GHz processors, 8 threads and 64GB memory with GTX1060 6GB GPU solely for feature matching and pruning. We conduct rotation averaging according to~\cite{chatterjee2013efficient}.  

{\it Methods and Datasets} We compare our proposed method with state-of the art BA-based approaches including ORB-SLAM~\cite{Mur-Artal2015a}, DVO-SLAM~\cite{kerl2013dense}, Kintinuous~\cite{Whelan12rssw}, LSD-SLAM~\cite{engel2014lsd} ElasticFusion~\cite{whelan2015elasticfusion}, ENFT-SFM~\cite{Zhang2016a}, VisualSfM~\cite{wu2011visualsfm}, LDSO~\cite{gao2018ldso} and two latest rotation-averaging-based SLAM systems: L-infinity SLAM~\cite{bustos2019} and Hybrid Camera Pose Estimation(HCPE)~\cite{Li20Hybrid}. Both of the hybrid approaches combine rotation averaging and canonical BA (re-projection error minimization). The approaches are tested on KITTI Odometry~\cite{Geiger2013c} and TUM-RGBD~\cite{sturm12iros} datasets with regards to both accuracy and backend processing time. In the experiments we use ORB features~\cite{rublee2011orb} for the frontend processing, followed by RANSAC~\cite{fischler1981random} iterations after feature matching to remove outliers. We also include an ablation study on how our 6-point feature selection affects the performance and show that ORB feature~\cite{rublee2011orb} combined with region selection achieves the better trade-off between accuracy and speed compared with brute-force feature matching.  
All the optimization steps are modified from~\cite{kummerle2011g, ceres-solver,l1magic} in C++. Note that all the recorded time is CPU processing time, we only use GPU for feature correspondence computation. 

\subsection{Evaluations on Real-World Benchmarks}

\begin{table}[h!]
\begin{center}
%\small
\resizebox{\linewidth}{!}{
\begin{tabular}%{@{\hspace{0.75mm}}r@{\hspace{0.95mm}}|@{\hspace{0.75mm}}c@{\hspace{.75mm}} @{\hspace{0.75mm}}c@{\hspace{.75mm}} @{\hspace{0.275mm}}c@{\hspace{.275mm}} @{\hspace{0.75mm}}c@{\hspace{.75mm}} | @{\hspace{0.75mm}}c@{\hspace{0.75mm}} @{\hspace{0.75mm}}c@{\hspace{0.75mm}} @{\hspace{0.75mm}}c@{\hspace{0.75mm}} @{\hspace{0.75mm}}c@{\hspace{0.275mm}}}% rcl  rcl rcl}
{r| c c c |c c c c }
\hline
 & {\thead{ENFT\\ SFM\\ \cite{Zhang2016a}}} & \thead{Visual\\ SFM\\ \cite{wu2011visualsfm}} &\thead{LDSO\\ \cite{gao2018ldso}} &\thead{ORB\\ SLAM\\ \cite{Mur-Artal2015a}} & \thead{L-infinity\\SLAM\\ \cite{bustos2019}}  & \thead{HCPE\\~\cite{Li20Hybrid}} & \thead{OURS} \\
\hline \hline
00&4.8&2.8 (3.7\%)&9.3&5.3 /15.26s & 6.7 /5.65s& \textcolor{red}{\textbf{4.9}} /3.42s & \textcolor{cyan}{\textbf{5.2}}/\textcolor{red}{\textbf{2.79s}}\\
\hline
01&57.2 &52.3 (12.5\%)& \textcolor{red}{\textbf{11.7}} & -  & 73.9 /9.23s & 45.9 /6.77s &\textcolor{red}{\textbf{42.6}} /\textcolor{red}{\textbf{4.35s}} \\
\hline
02&28.3&1.8 (4.5\%)& 32.0&21.3 /21.4s&27.7/3.90s & \textcolor{red}{\textbf{18.2}}/4.43s &\textcolor{cyan}{\textbf{18.7}}/\textcolor{red}{\textbf{3.63s}}\\
\hline
03&2.9&0.3 (12.0\%)&2.9 &\textcolor{red}{\textbf{1.5}} /1.97s& 8.2/1.01s& \textcolor{red}{\textbf{1.5}}/0.78s &\textcolor{cyan}{\textbf{2.2}}/\textcolor{red}{\textbf{0.73s}}\\
\hline
04&0.7&0.8 (23.4\%)&1.2&1.6 /1.02s&1.4/0.78s & \textcolor{red}{\textbf{0.9}}/0.14s & \textcolor{red}{\textbf{0.9}}/\textcolor{red}{\textbf{0.13s}}\\
\hline
05&3.5&9.8 (7.4\%)&5.1&2.9 /6.93s&5.7/6.14s &\textcolor{cyan}{\textbf{2.9}}/4.87s &\textcolor{red}{\textbf{2.7}}/\textcolor{red}{\textbf{3.99s}}\\
\hline
06&14.4&8.6 (7.4\%)&13.6&12.3 /2.87s& 21.3/1.77s& \textcolor{red}{\textbf{8.2}}/0.65s &\textcolor{cyan}{\textbf{8.7}}/\textcolor{red}{\textbf{0.62s}}\\
\hline
        07&\textcolor{red}{\textbf{2.0}}&3.9 (7.8\%)&3.0&\textcolor{cyan}{\textbf{2.3}} /1.22s&2.9/1.22s &\textcolor{red}{\textbf{2.2}}/\textcolor{red}{\textbf{0.70s}} & 2.5/0.72s\\
\hline
08&28.3&0.8 (0.9\%)&129.0&46.7 /20.9s&- &\textcolor{red}{\textbf{33.6}}/10.09s& \textcolor{cyan}{\textbf{42.4}}/\textcolor{red}{\textbf{8.23s}}\\
\hline
09&\textcolor{red}{\textbf{5.9}}&0.9 (4.9\%)&21.7&6.6 /4.47s&9.2/3.71s & \textcolor{red}{\textbf{6.5}}/1.96s &\textcolor{cyan}{\textbf{7.2}}/\textcolor{red}{\textbf{1.66s}}\\
\hline
10&18.5&5.7 (6.5\%)&17.4&8.8 /3.72s&21.4/1.23s & \textcolor{red}{\textbf{7.6}}/\textcolor{red}{\textbf{0.69s}} &\textcolor{cyan}{\textbf{8.2}}/1.30s\\
\hline
\end{tabular}}
\caption{\small Trajectory estimation RMSE(m) and pose estimation runtime on KITTI Odometry. For VisualSFM the number in the parenthesis represents the map completeness ratio and it is 100\% for all other methods. For the last four approaches including ours, the second number represents the back-end optimization runtime. Note that ENFT-SFM and VisualSFM are offline SfM systems and the latter requires preprocessings. ORB-SLAM fails to initialize on Seq.01 so we leave the result blank. We use the same set of optimization parameters for our approach in all the experiments.}
\label{table:kittiquant}
\end{center}
%\vspace{-.1cm}
\end{table}

\textbf{Evaluation on KITTI Odometry} The KITTI dataset contains 11 video sequences captured by a fast-moving vehicle over long trajectories. We average the results over 20 runs and record the trajectory estimation error and backend optimization time in Table.~\ref{table:kittiquant}. For sequential data, rotation averaging based approaches achieve clearly faster processing time while ORB-SLAM fails to process Seq. 01, where repeated failures of initialization is observed.

\begin{figure}[!ht]
\begin{center}
\subfloat	[Sequence 00]	{
\includegraphics[width=.5\linewidth]{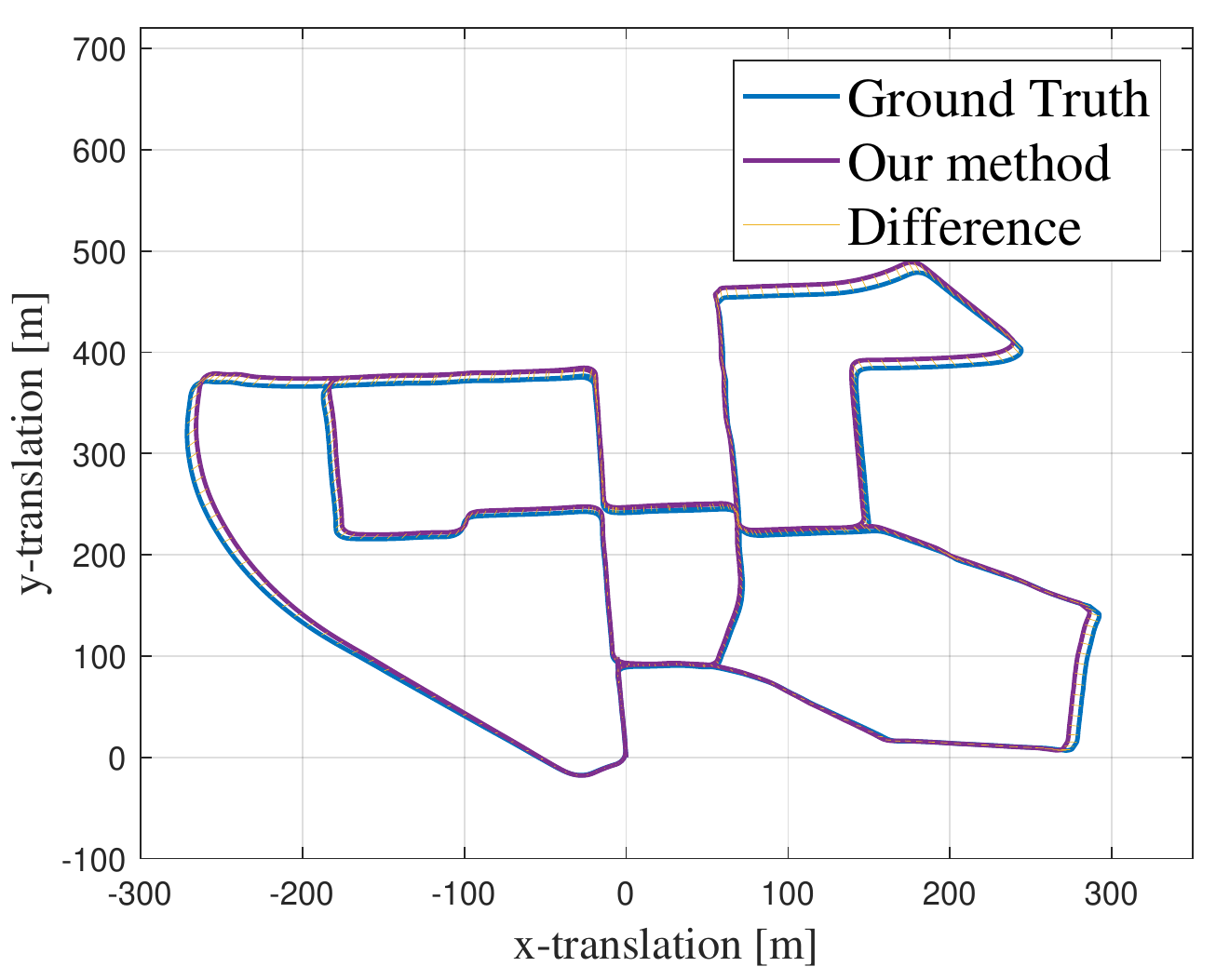}
}
\subfloat	[Sequence 02]	{
\includegraphics[width=.5\linewidth]{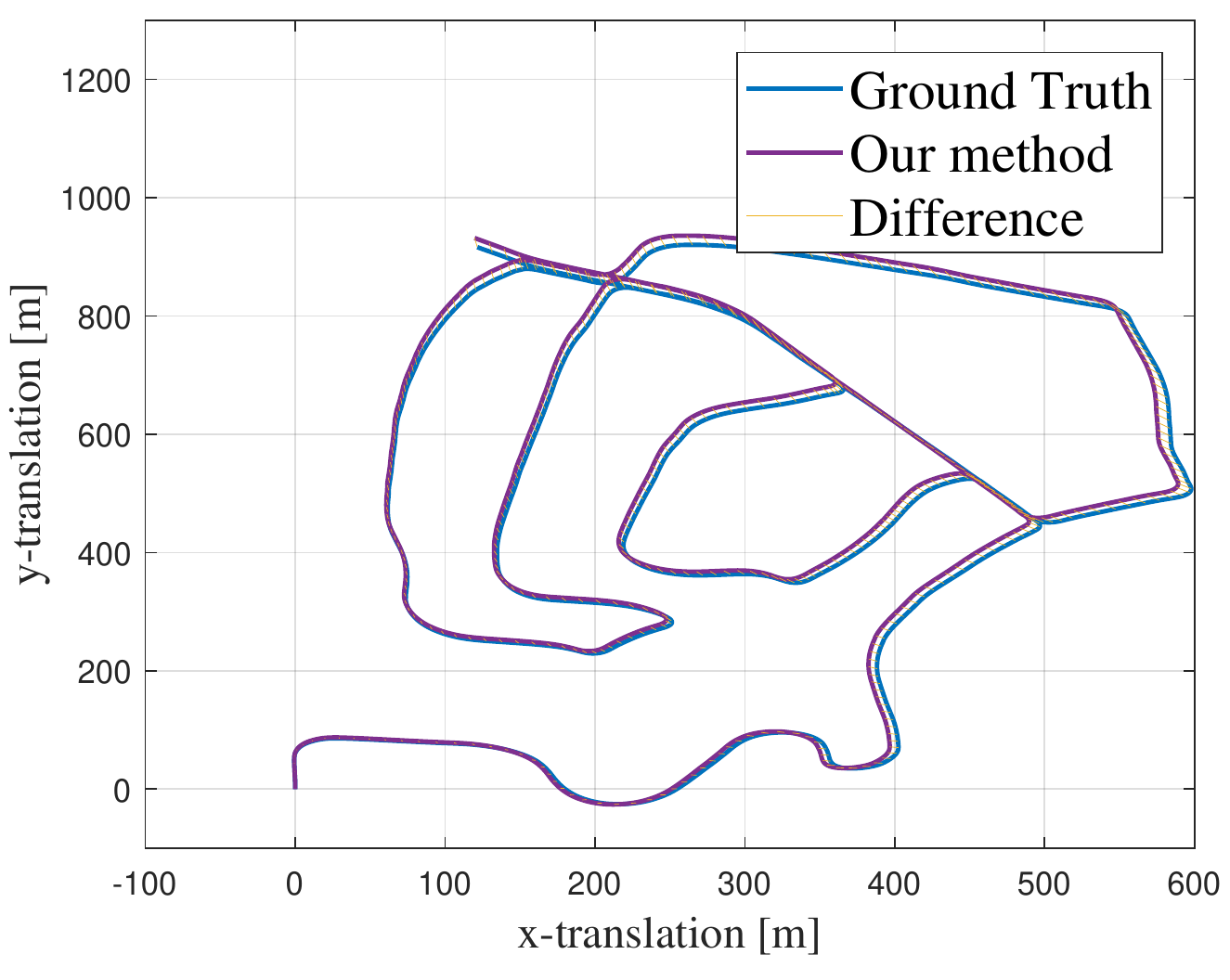}
}

\subfloat  [Sequence 05]    {
\includegraphics[width=0.5\linewidth]{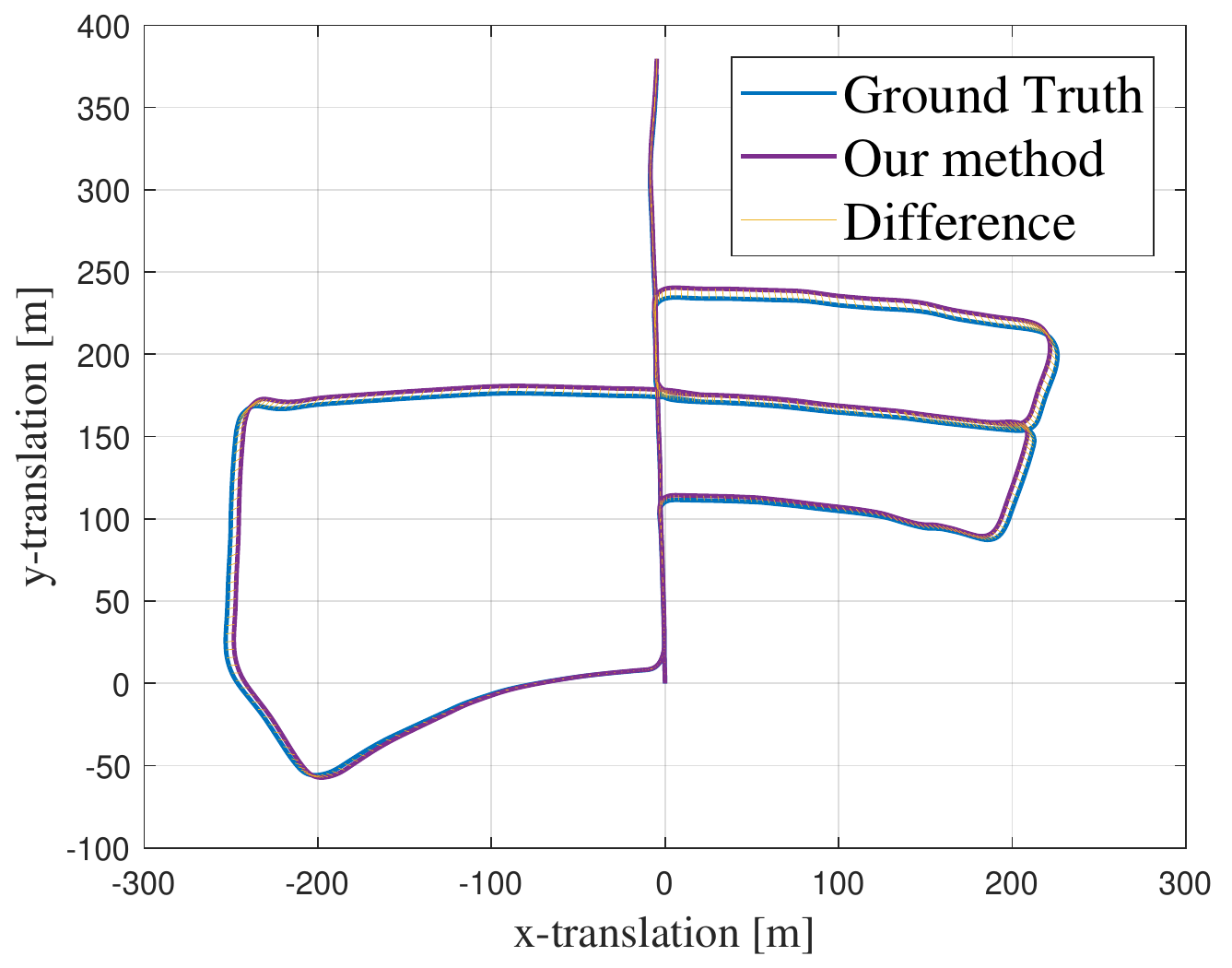}
}
%%\vspace{-1pt}
\subfloat   [Sequence 09]   {
\includegraphics[width=0.5\linewidth]{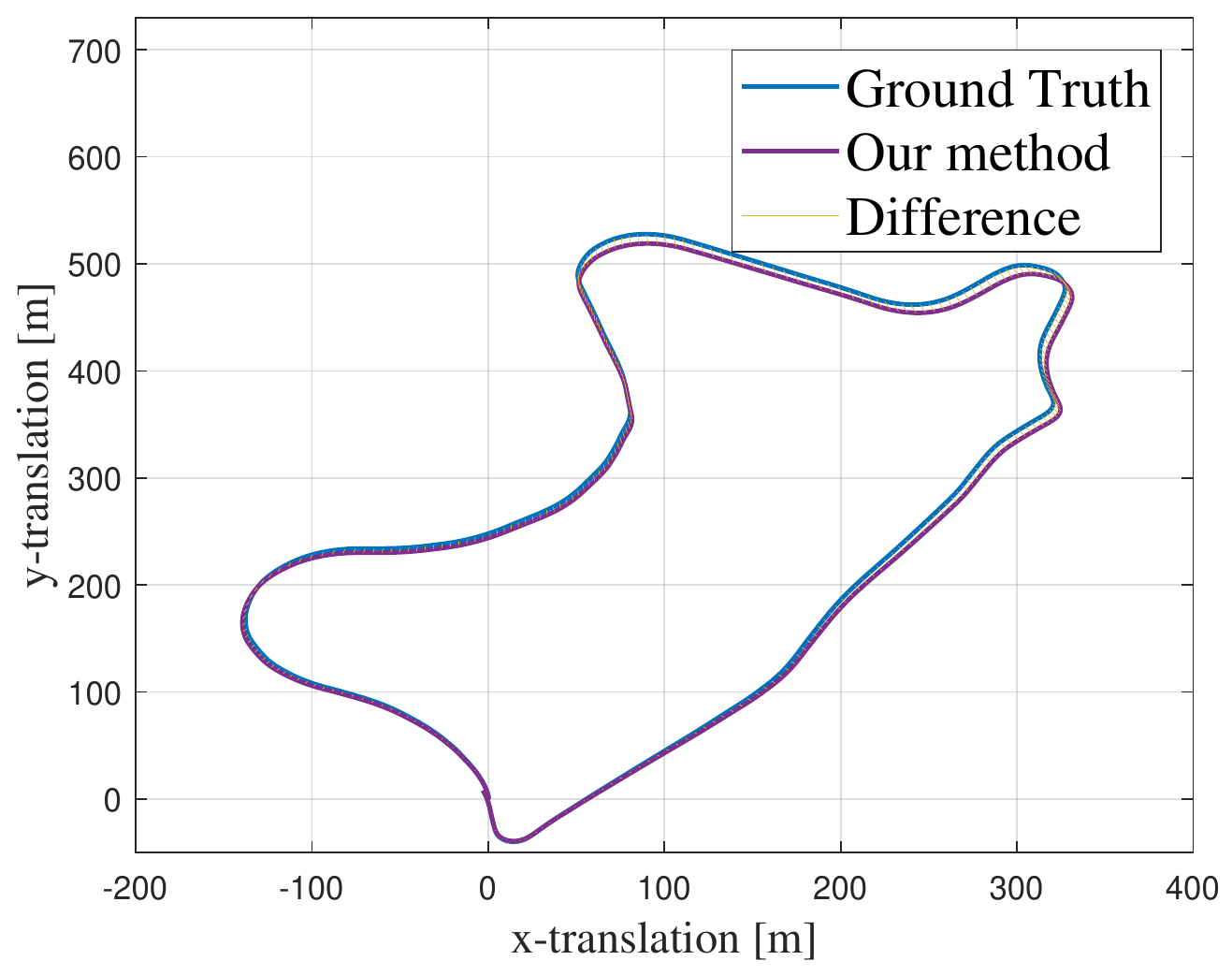}
}
%%\vspace{-0.1cm}
\caption[Comparison with ORB-SLAM on Seq.02, 08, 10 from KITTI Odometry.]{Our proposed approach shows a trajectory estimation of comparably high quality in KITTI dataset which contains large scale outdoor scenes. Specifically, our approach displays superior loop-closing capability on Seq.02, where many loops are present. It can be shown that the accumulated drifts are negligible through the trajectory estimation. Results on the rest of the dataset are provided in the supplementary materials.}
\label{fig:3seq}
\end{center}
%\vspace{-.3cm}
\end{figure}

\begin{figure*}[!ht]
\setlength{\belowcaptionskip}{-20pt}
\begin{center}
\subfloat	[freiburg1\_desk]	{
\includegraphics[width=.315\linewidth]{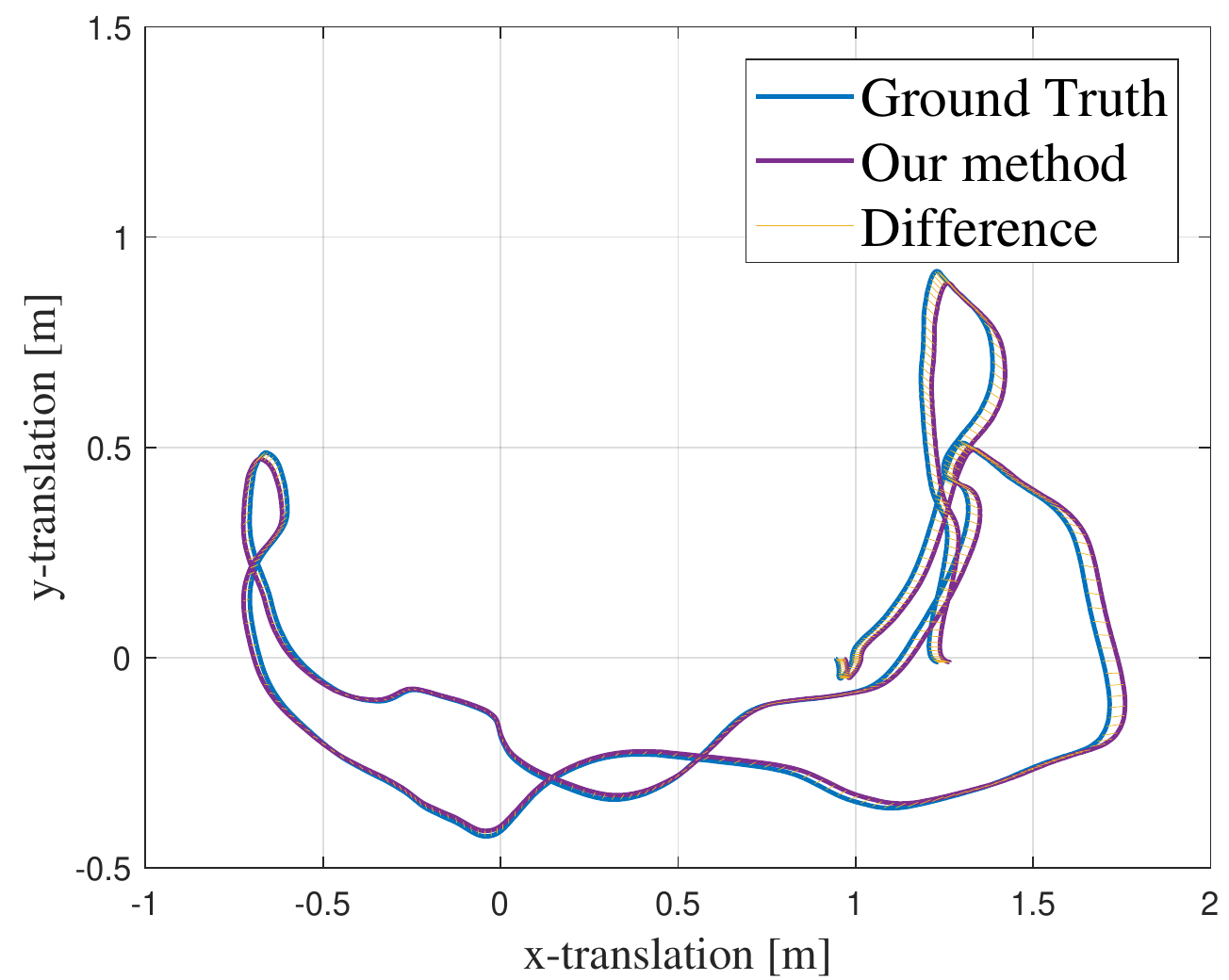}
}
\subfloat  [freiburg1\_room]    {
\includegraphics[width=.315\linewidth]{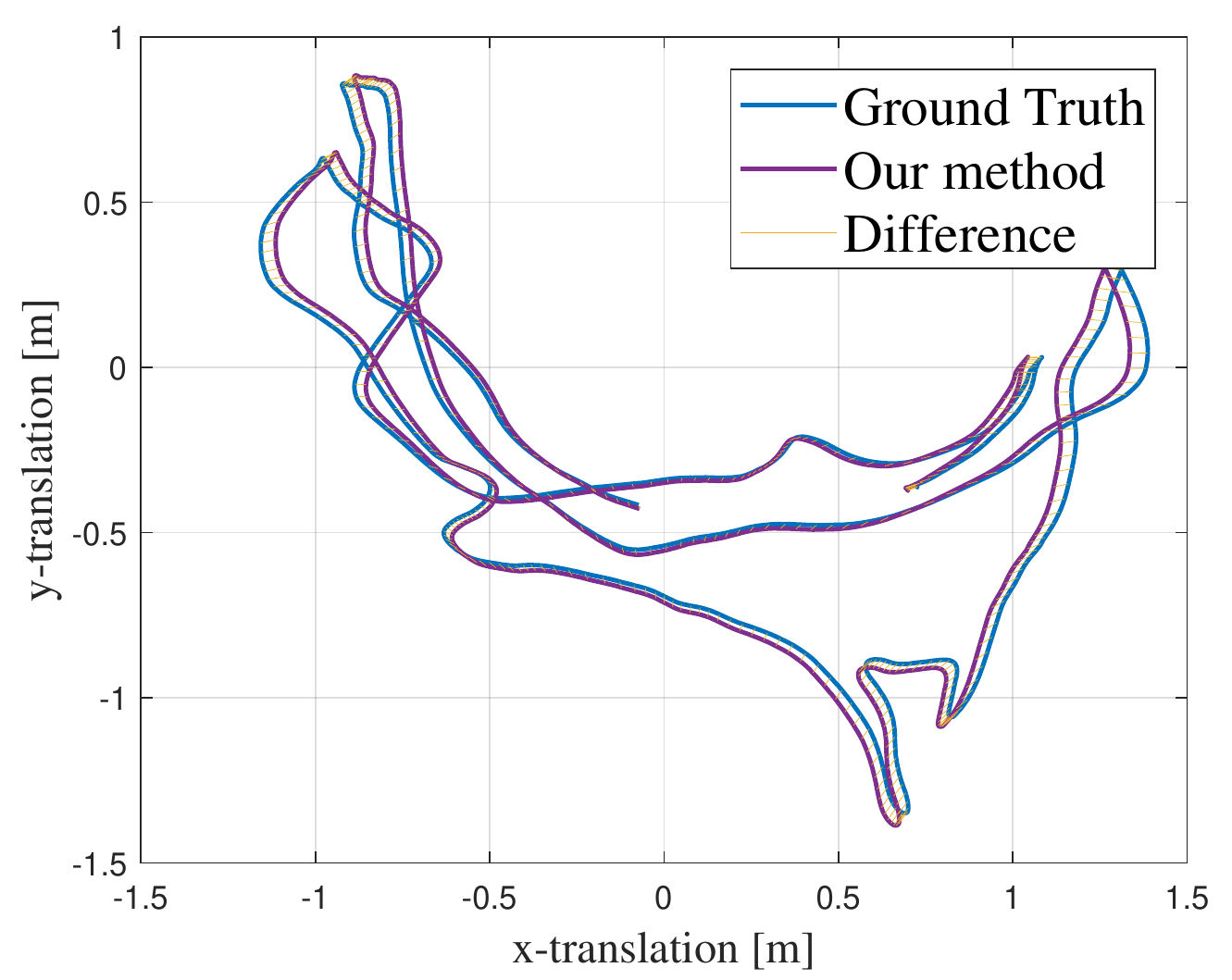}
}
%%\vspace{-1pt}
\subfloat   [freiburg3\_long\_office\_household]   {
\includegraphics[width=.3\linewidth]{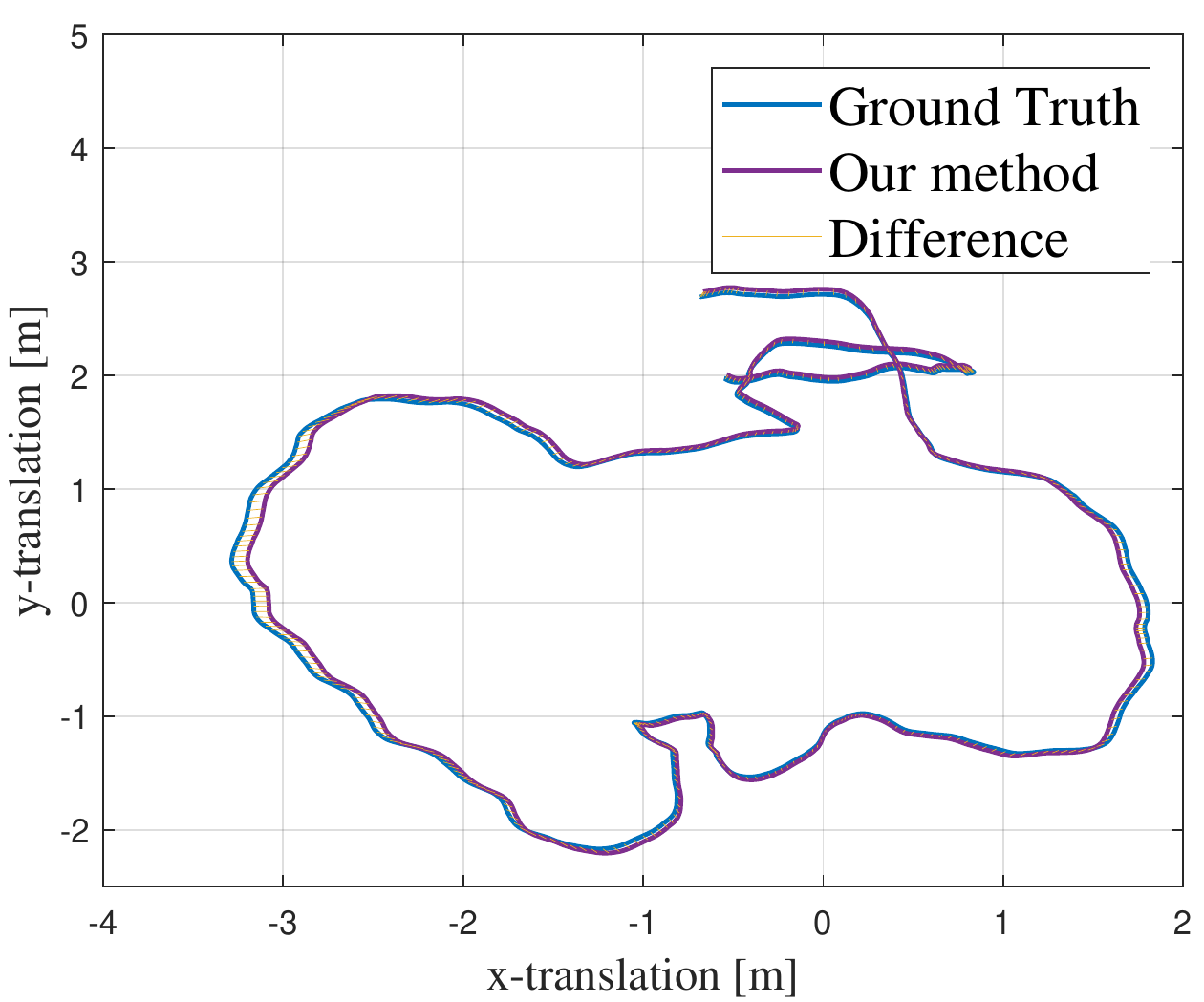}
}
%%\vspace{-0.1cm}
\caption[Comparison with ORB-SLAM on Seq.02, 08, 10 from KITTI Odometry.]{{\textbf TUM-RGBD} Our proposed approach shows accurate trajectory estimation in small-scale indoor scenes where many challenges are present like light variations, textureless surface scanning, pure rotational motions. In all the sequences it can be shown that there is not many drift accumulation. More clear figures are provided in supplementary materials.}
\label{fig:3seq}
\end{center}
%\vspace{-.3cm}
\end{figure*}

Moreover, in the experiments we notice that L-infinity SLAM is much more sensitive to outliers due to the $l_\infty$ optimization. Hence for KITTI dataset where many repetitive patterns are present, L-infinity SLAM achieves the lowerst accuracy on all of the sequences. Also L-infinity SLAM fails on Seq. 08, we conjecture the reason might be that the system fails on loop closing due to noisy features, and the accumulated drift is exaggerated within $l_1$-norm framework, winding up the eventual loss of tracking.

Our approach achieves the fastest processing time on most of the sequences. The accuracy is not as high as HCPE since our proposed approach is processed in a keyframed pipeline where only temporal covisibility is involved while HCPE utilized dense data association, yielding a slightly higher accuracy.

\textbf{Evaluation on TUM-RGBD dataset} The TUM RGB-D dataset contains diverse indoor video sequences, including many challenging scenes, \eg, rotation-only motion, textureless surface scanning in a small-scale scene. In all of our experiments, we only use RGB data without relying on the depth data. We present selected experiment results in Table.~\ref{table:tum}. Among the state-of-the-art approaches, DVO-SLAM, Kintinuous and Elastic Fusion use the depth data to run the experiments. Also, ENFT-SFM and VisualSfM are both offline systems, where experiment results of VisualSfM rely on manually-selected keyframes. Qualitative results on 3 of the sequences are given in Fig.~\ref{fig:3seq}.

\begin{table}[!t]
\setlength{\belowcaptionskip}{-10pt}
\begin{center}
\resizebox{\linewidth}{!}{%
%\centering
%\scriptsize
\begin{tabular}{r | c c c c c c c c}
\hline
& \thead{DVO\\SLAM \\~\cite{kerl2013dense}} & \thead{Kinti-\\nuous\\~\cite{Whelan12rssw}} &\thead{LSD\\ SLAM\\~\cite{engel2014lsd} } &\thead{Elastic\\ Fusion\\~\cite{whelan2015elasticfusion}} & \thead{ORB\\SLAM\\~\cite{Mur-Artal2015a}}& \thead{L-infinity\\SLAM\\~\cite{bustos2019}} &\thead{HPCE\\~\cite{Li20Hybrid}}&\thead{OURS\\ }\\
\hline \hline 
\footnotesize{fr1\_desk} & 2.1 &3.7&10.65&2.0&1.7&3.2*&\textcolor{cyan}{\textbf{1.6}}&\textcolor{red}{\textbf{1.2}}\\
\hline
\small{fr1\_desk2} & 4.6 & 7.1&47.6&4.8&2.9*&4.6*&\textcolor{red}{\textbf{1.5}}&\textcolor{cyan}{\textbf{2.2}}\\
\hline
\small{fr1\_room} & 4.3 & 7.5&39.2&6.8&3.4*&3.7*&\textcolor{red}{\textbf{2.9}}&\textcolor{cyan}{\textbf{3.2}}\\
\hline
\small{fr2\_desk} & 1.7 & 3.4 & 4.57 & 7.1&\textcolor{red}{\textbf{0.8}}&2.3*&1.5&\textcolor{cyan}{\textbf{0.9}}\\
\hline
\small{fr2\_xyz} & 1.8&2.9&6.29&1.1&\textcolor{red}{\textbf{0.3}}&1.2*&\textcolor{red}{\textbf{0.3}}&\textcolor{cyan}{\textbf{0.4}}\\
\hline
\small{fr3\_long\_office} & 3.5 & 3.0 & 38.53& 1.7&3.5&2.9*&\textcolor{red}{\textbf{0.8}}&\textcolor{red}{\textbf{0.8}}\\
\hline
\small{fr3\_nst\_near} & 1.8 & 3.1& 7.54& 1.6&{1.4}&2.2*&\textcolor{cyan}{\textbf{1.1}}&\textcolor{red}{\textbf{0.9}}\\
\hline
\end{tabular}}
\caption{\small Trajectory estimation RMSE (cm) on the TUM RGB-D \cite{sturm12iros} dataset. Results with (*) are experimental results based on 5-execution medians from running open-source codes provided by the authors. The first five systems are the BA-based state of the art while the last three approaches including ours primarily rely on rotation averaging.}
\label{table:tum}
\end{center}
%\vspace{-.5cm}
\end{table}
\subsection{Ablation Study on Feature Selection}
In addition, we conduct the ablation study on how feature point selection affects the SLAM accuracy. We run the experiment on KITTI Seq.02 with three different feature selection schemes. For the random selection of 6 points, the 6 feature correspondences are selected from the whole feature space without any constraints; For the region-of-interest(ROI) selection scheme, we first process feature clustering on the feature space such that different regions are sorted according to the amounts of the feature points, then the 6 points are all selected from the highest-score ROI. It is clearly shown that our proposed point selection performs superior to the other two approaches on accuracy.
\begin{figure}
    \centering
    \includegraphics[width=0.8\linewidth]{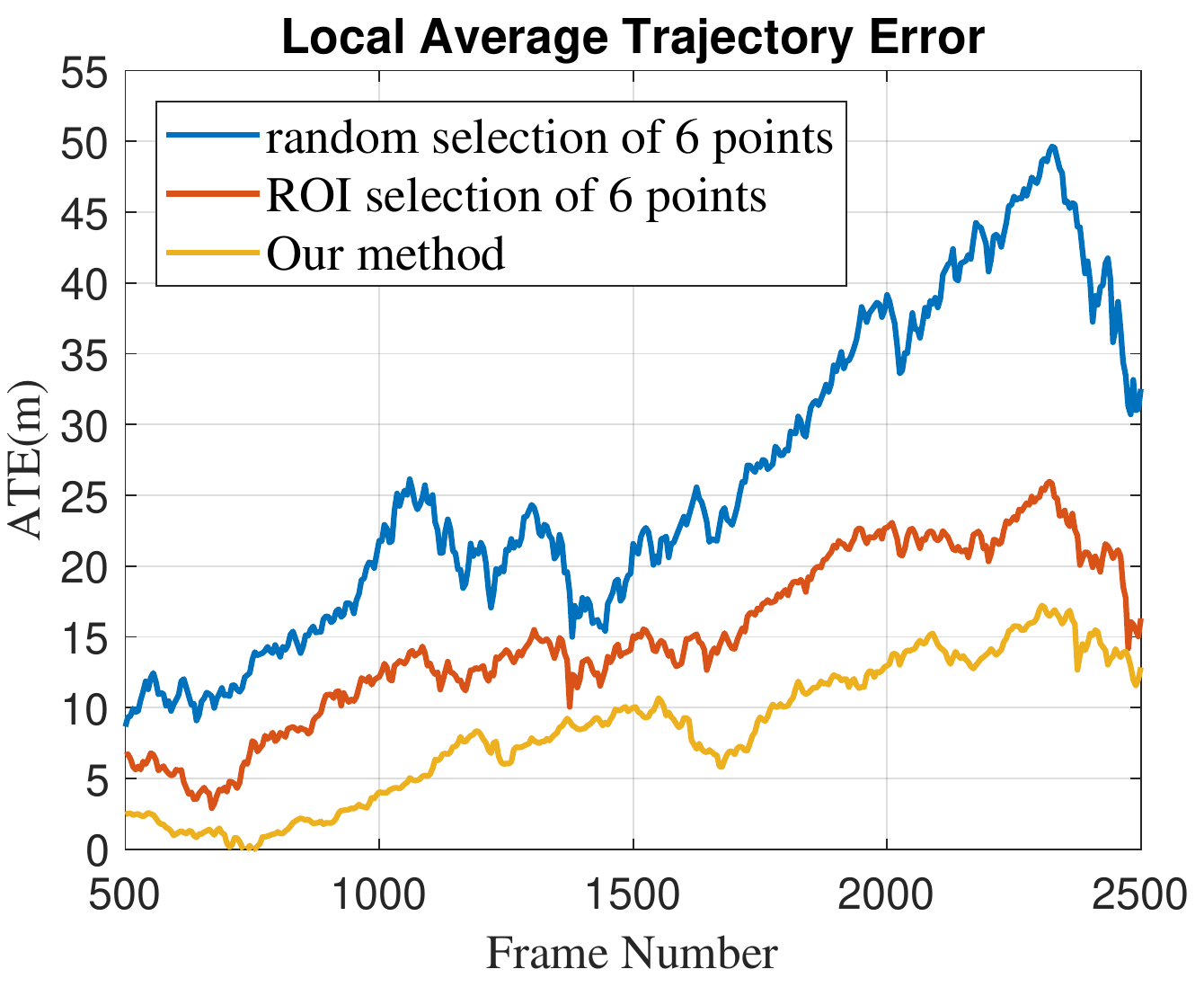}
    \caption{We repeat the experiments with three different point selection schemes. While ROI selection performs slightly better than the random feature selection, both schemes yield much higher trajectory estimation error than our proposed approach.}
    \label{fig:my_label}
\end{figure}

\section{Conclusion}
In this paper, we propose a rotation averaging optimization framework for backend of visual SLAM systems. With decoupling camera rotation and translation, we utilize epipolar geometric constraints to estimate the relative camera rotations, followed by state-of-the-art rotation averaging method to obtain the absolute camera rotation in global coordinate system. Absolute camera translations are then recovered by minimizing geometric errors, which is more efficient than utilizing point-camera correspondence. We further analyze the global optimality of our solution and address critical problem in monocular SLAM, including initialization and loop closure. Experiment results further validates the efficiency and robustness of our approach.

\clearpage
%\newpage
{
%\nocite{*}
\bibliographystyle{ieee_fullname}
\bibliography{egapaper}
}
\clearpage
%\section{KITTI}
%\section{Qualitative Results}
\begin{figure*}[h!]
%\caption{Comparison of Reconstructions}
\label{fig:kitti}
\begin{center}
\subfloat   [Sequence 00]  {
\includegraphics[width=.33\linewidth]{seq00.pdf}
}
\subfloat   [Sequence 01]  {
\includegraphics[width=.33\linewidth]{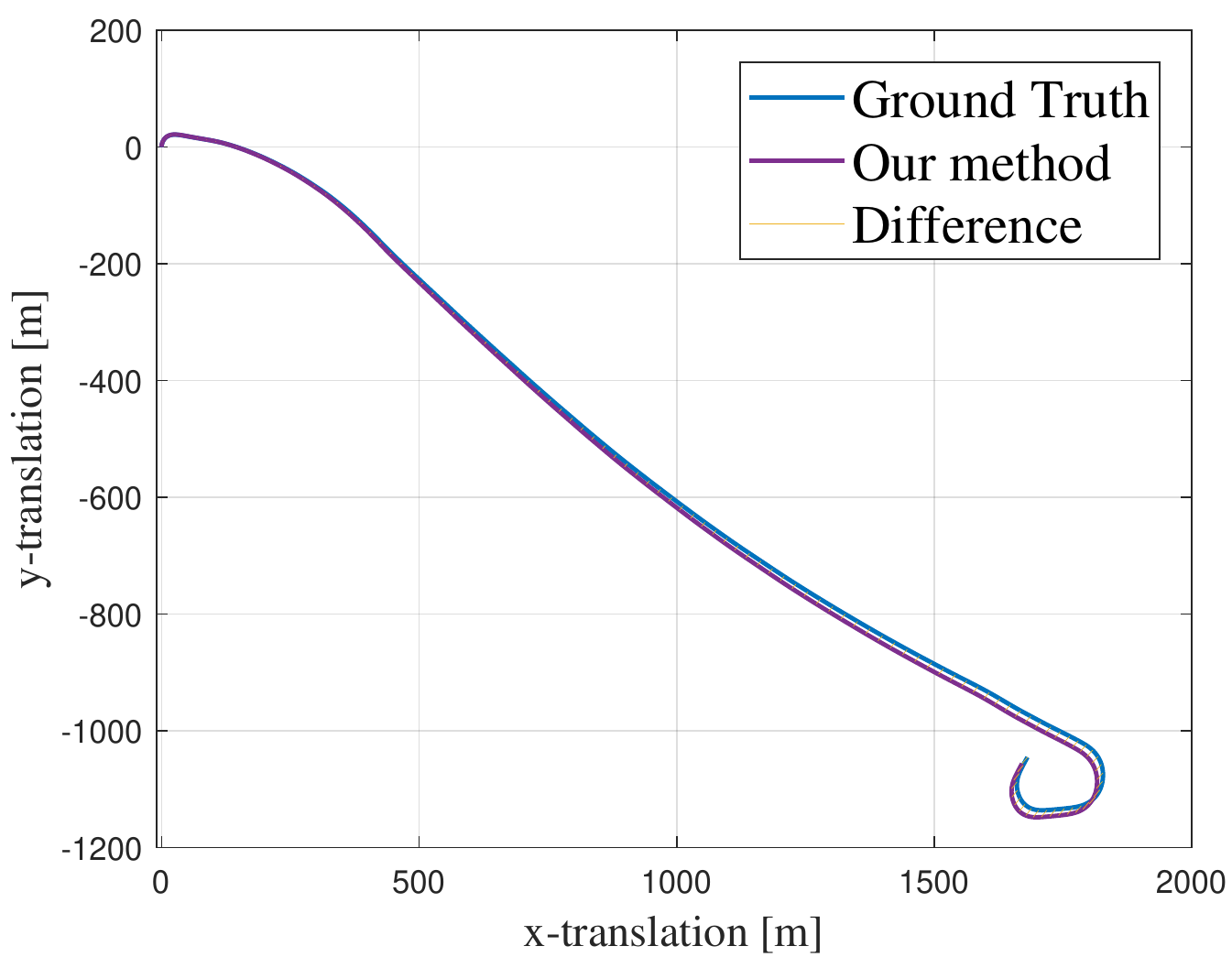}
}
%\vspace{-1mm}
\subfloat   [Sequence 03] {
\includegraphics[width=.33\linewidth]{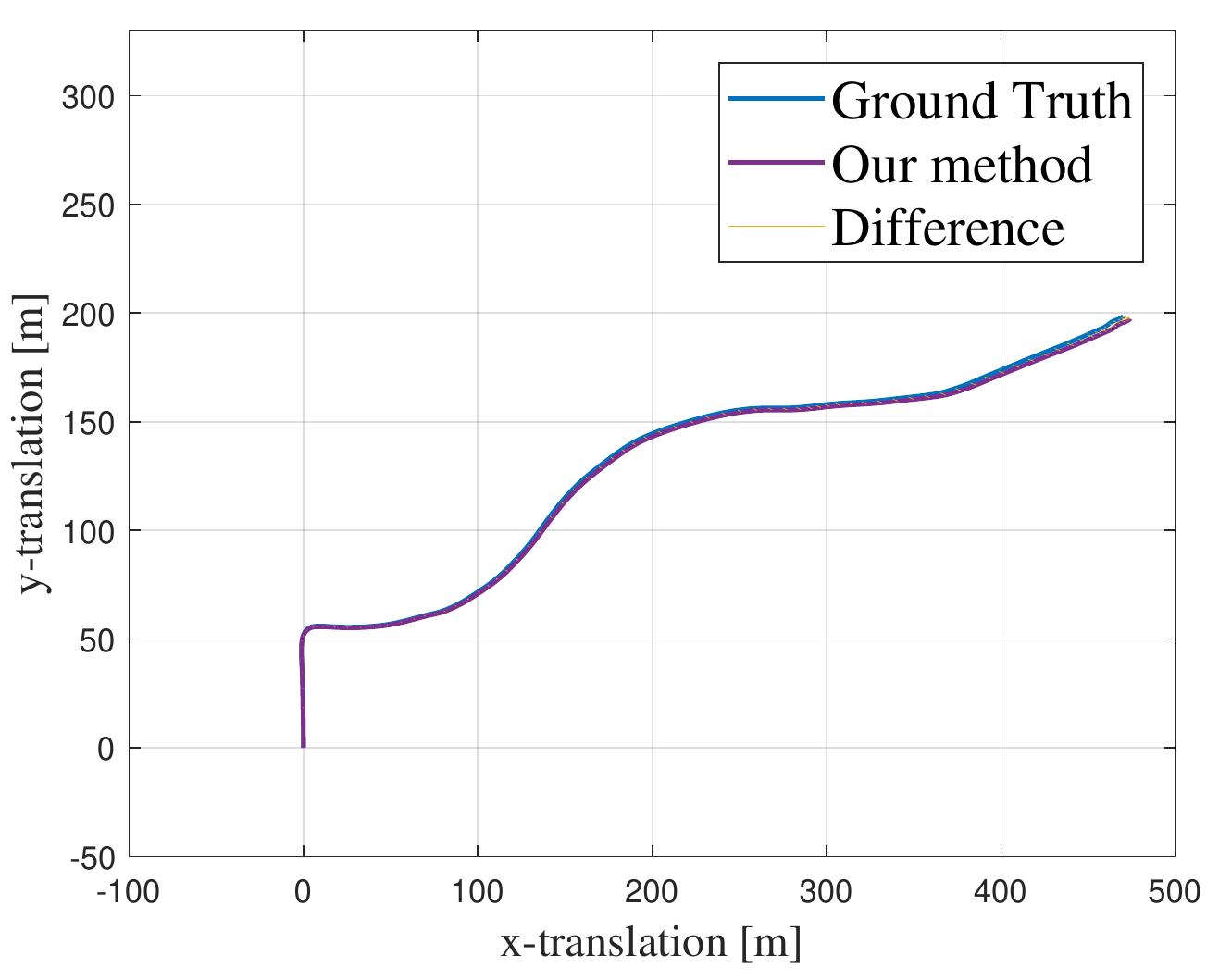}
}
\vspace{3mm}
\subfloat  [Sequence 04]    {
\includegraphics[width=.33\linewidth]{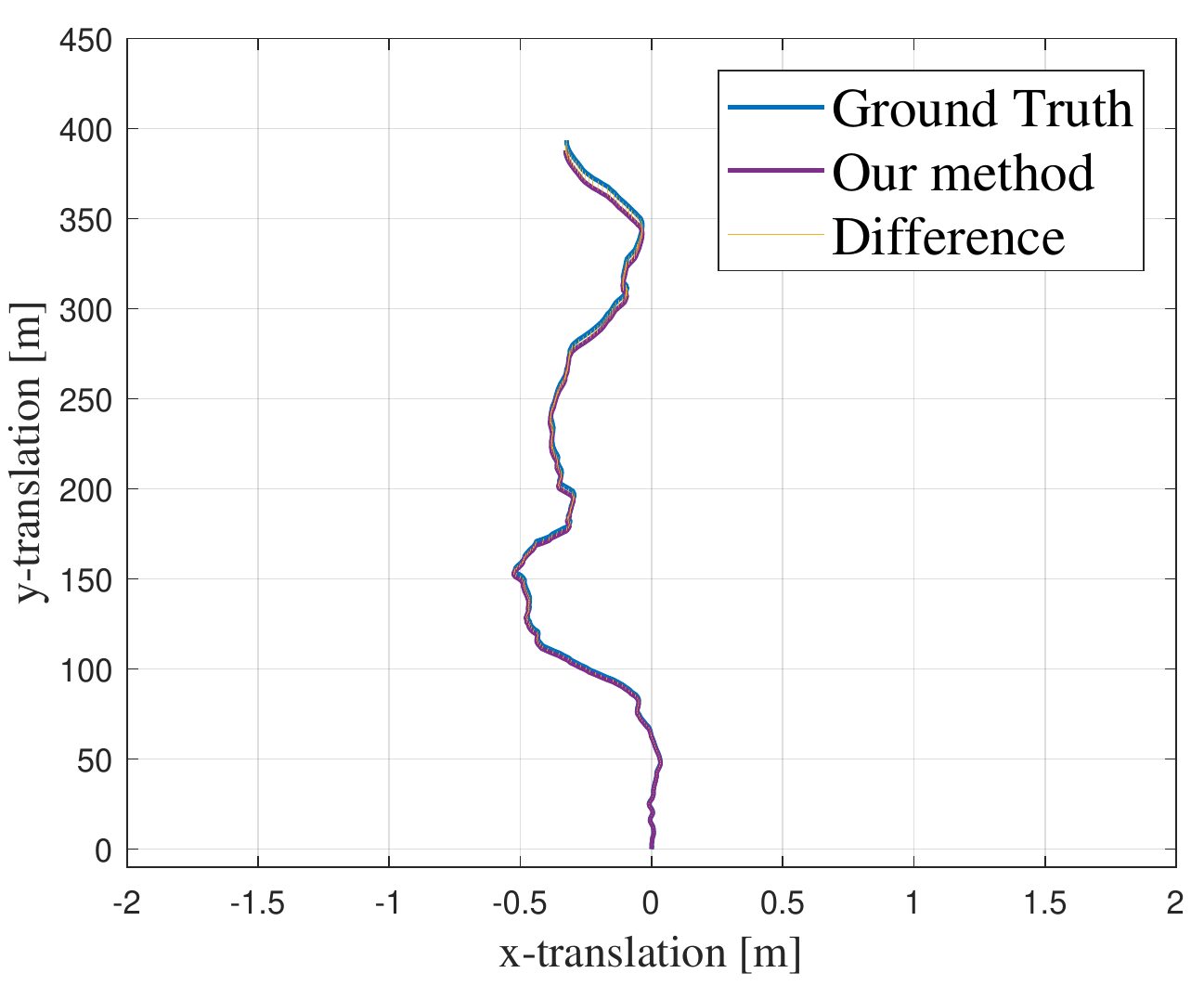}
}
\subfloat   [Sequence 05]   {
\includegraphics[width=.33\linewidth]{seq05.pdf}
}
%\vspace{-0.1cm}
\subfloat   [Sequence 06]   {
\includegraphics[width=.33\linewidth]{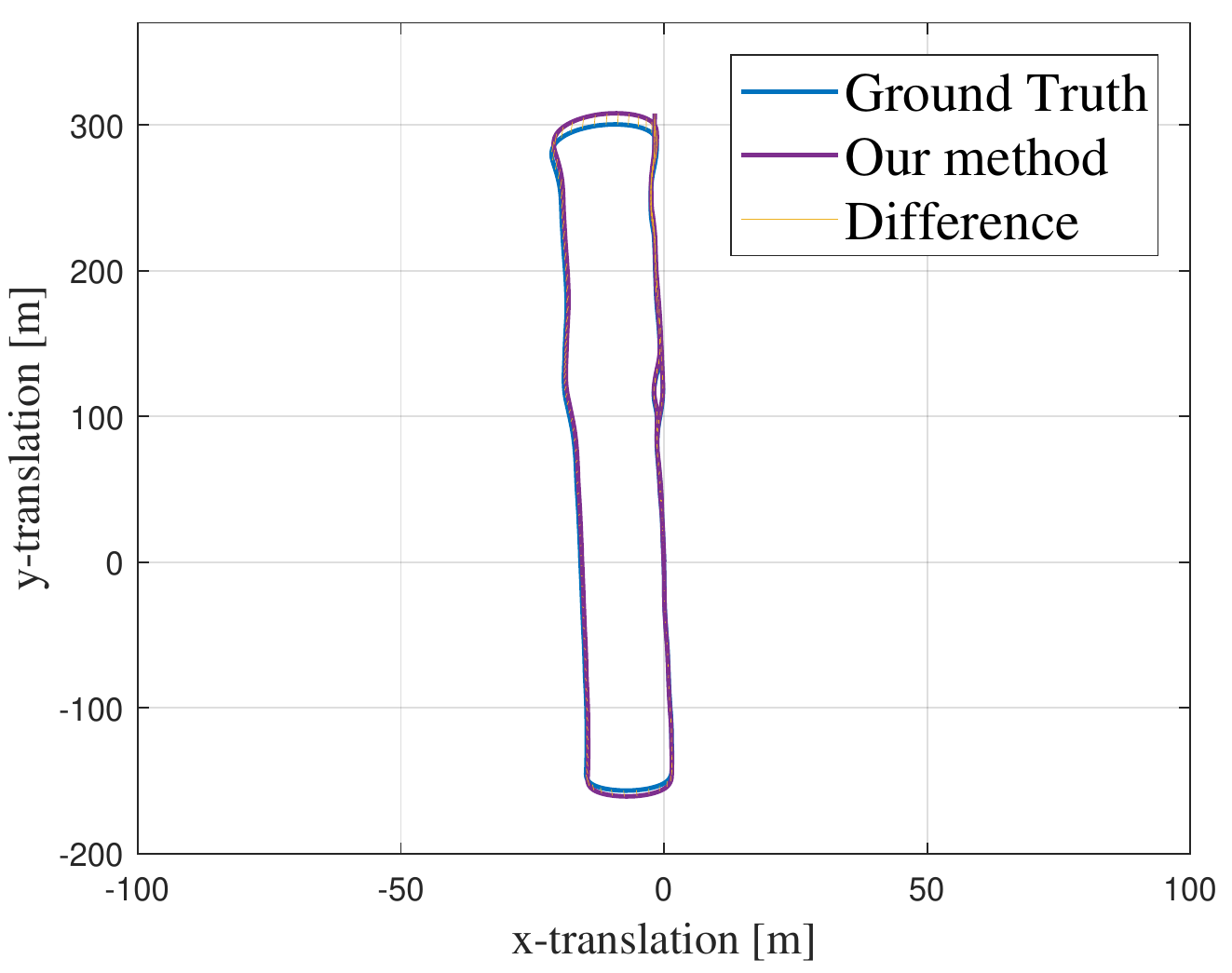}
}
\vspace{3mm}
\subfloat    [Sequence 07]  {
\includegraphics[width=.33\linewidth]{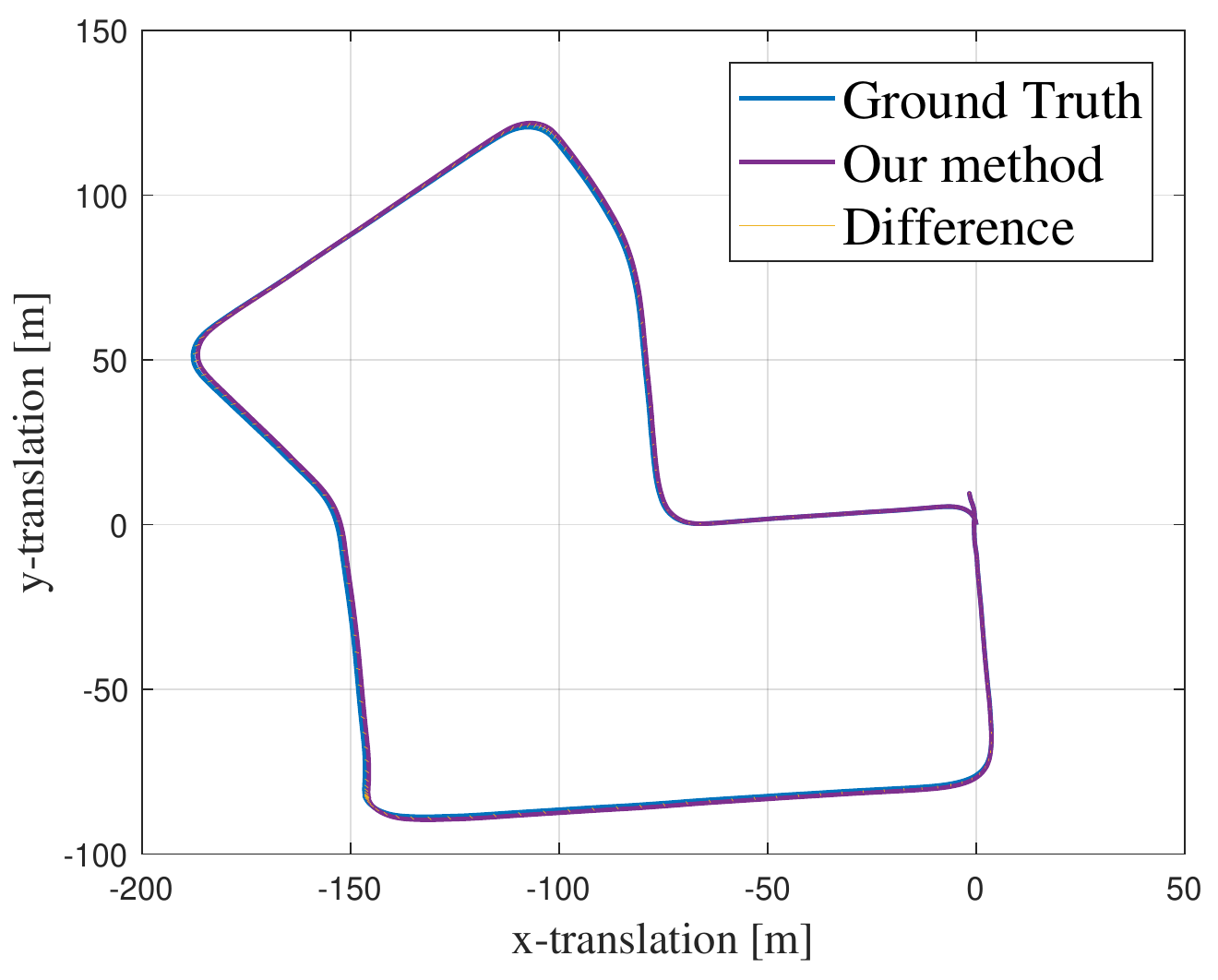}
}
\subfloat   [Sequence 09]   {
\includegraphics[width=.33\linewidth]{seq09.pdf}
}
\caption{\small Trajectories of our proposed system on the KITTI dataset.}
\label{fig:kitti}
\vspace{-2.11mm}
\end{center}
\end{figure*}

\begin{figure*}[!ht]
\setlength{\belowcaptionskip}{-20pt}
\begin{center}
\subfloat	[freiburg1\_desk]	{
\includegraphics[width=.47\linewidth]{f1desk1.pdf}
}
\subfloat  [freiburg1\_room]    {
\includegraphics[width=.47\linewidth]{fr1room.pdf}
}
%%\vspace{-1pt}

\subfloat   [freiburg3\_long\_office\_household]   {
\includegraphics[width=.47\linewidth]{f3longoffice.pdf}
}
\subfloat   [freiburg3\_floor]   {
\includegraphics[width=.49\linewidth]{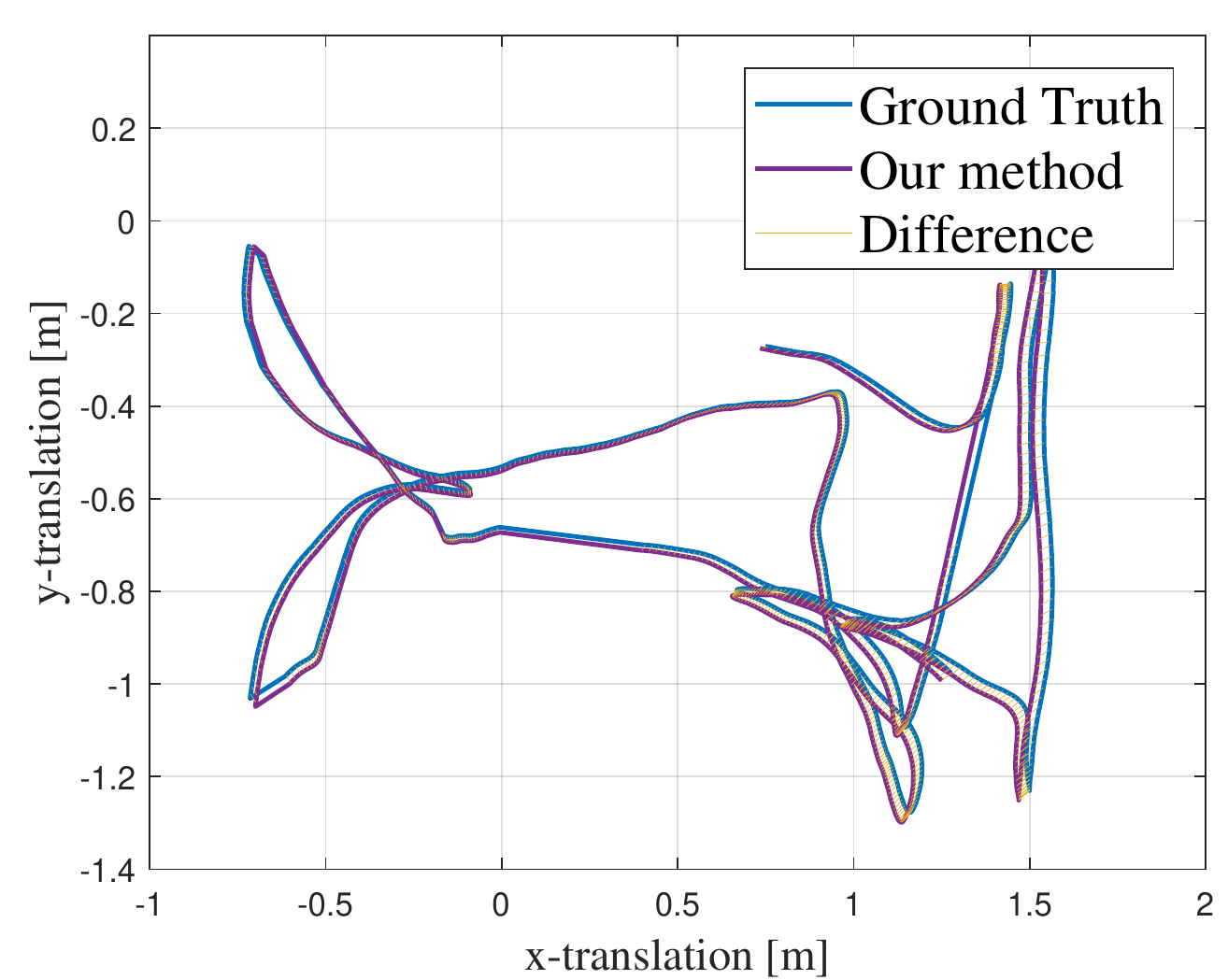}
}
%\setlength{\belowcaptionskip}{-10pt}
%%\vspace{-0.1cm}
\caption[Comparison with ORB-SLAM on Seq.02, 08, 10 from KITTI Odometry.]{Trajectories of our proposed system on the TUM RGB-D dataset.}
\label{fig:3seq}
\end{center}
%\vspace{-.3cm}
\end{figure*}

\end{document}